\pdfoutput=1

\documentclass[11pt]{article}

\usepackage[]{ACL2023}

\usepackage{times}
\usepackage{latexsym}

\usepackage[T1]{fontenc}

\usepackage[utf8]{inputenc}

\usepackage{microtype}

\usepackage{inconsolata}

\usepackage{CJKutf8}
\usepackage{CJKpunct}
\usepackage{CJKnumb}

\usepackage{times}
\usepackage{soul}
\usepackage{url}
\usepackage{latexsym}
\usepackage{graphicx}
\usepackage{verbatim}
\usepackage{booktabs}
\usepackage{amssymb}
\usepackage{stfloats}
\usepackage{array}
\usepackage{bm}
\usepackage{amsfonts}
\usepackage{amsmath}
\usepackage{multirow}
\usepackage{bbding}
\usepackage{amsthm}
\usepackage{algorithm}
\usepackage{algorithmic}
\usepackage{comment}
\usepackage{color}
\usepackage{makecell}
\usepackage{subcaption}
\usepackage{longtable}

\definecolor{darkgreen}{RGB}{67,182,84}

\title{An Extensible Plug-and-Play Method for Multi-Aspect \\
 Controllable Text Generation}

\author{
    Xuancheng Huang$^{1\dag}$ \ 
    Zijun Liu$^{2,4\dag}$ \ 
    Peng Li$^{3,5}$ \ 
    Tao Li$^{1}$ \ 
    Maosong Sun$^{2,4*}$ \ 
    Yang Liu$^{2,3,4,5*}$ \\
    $^1$Meituan, China \\
    $^2$Dept. of Comp. Sci. \& Tech., Institute for AI, Tsinghua University, Beijing, China \\
    $^3$Institute for AI Industry Research (AIR), Tsinghua University, Beijing, China \\
    $^4$Beijing National Research Center for Information Science and Technology \\
    $^5$Shanghai Artificial Intelligence Laboratory, Shanghai, China\\
}
\begin{document}
\maketitle

\begin{CJK*}{UTF8}{gbsn}

\begin{abstract}
{\let\thefootnote\relax\footnotetext{$\dag$ \ indicates equal contribution.}}
{\let\thefootnote\relax\footnotetext{*\ \ Corresponding authors: M.Sun (\href{mailto:sms@tsinghua.edu.cn}{\texttt{sms@tsinghua.edu}} \href{mailto:sms@tsinghua.edu.cn}{\texttt{.cn}}) and Y.Liu (\href{mailto:liuyang2011@tsinghua.edu.cn}{\texttt{liuyang2011@tsinghua.edu.cn}}) }}

Recently, multi-aspect controllable text generation that controls the generated text in multiple aspects (e.g., sentiment, topic, and keywords) has attracted increasing attention. Although methods based on parameter efficient tuning like prefix-tuning could achieve multi-aspect controlling in a plug-and-play way, the mutual interference of multiple prefixes leads to significant degeneration of constraints and limits their extensibility to training-time unseen aspect combinations. In this work, we provide a theoretical lower bound for the interference and empirically found that the interference grows with the number of layers where prefixes are inserted. Based on these analyses, we propose using trainable gates to normalize the intervention of prefixes to restrain the growing interference. As a result, controlling training-time unseen combinations of aspects can be realized by simply concatenating corresponding plugins such that new constraints can be extended at a lower cost. In addition, we propose a unified way to process both categorical and free-form constraints. Experiments on text generation and machine translation demonstrate the superiority of our approach over baselines on constraint accuracy, text quality, and extensibility.\footnote{The source code is available at \url{https://github.com/THUNLP-MT/PromptGating4MCTG}}

\end{abstract}

\section{Introduction}

Multi-aspect controllable text generation (MCTG), which aims at generating fluent text while satisfying multiple aspects of constraints simultaneously, has attracted increasing attention in recent years~\cite{DBLP:conf/iclr/ChanOPZF21,qian-etal-2022-controllable,DBLP:journals/corr/abs-2210-02889}. To effectively control diverse aspects such as sentiment, topic, and detoxification, extensive efforts have been devoted to the task, including methods based on conditional generative model~\cite{keskar2019ctrl}, decoding-time regulation~\cite{lin-riedl-2021-plug,DBLP:conf/nips/KumarMST21}, and parameter efficient tuning~\cite{qian-etal-2022-controllable,DBLP:journals/corr/abs-2210-02889}.

\begin{figure*}[!t]
  \centering
  \resizebox{\linewidth}{!}{
  \includegraphics[]{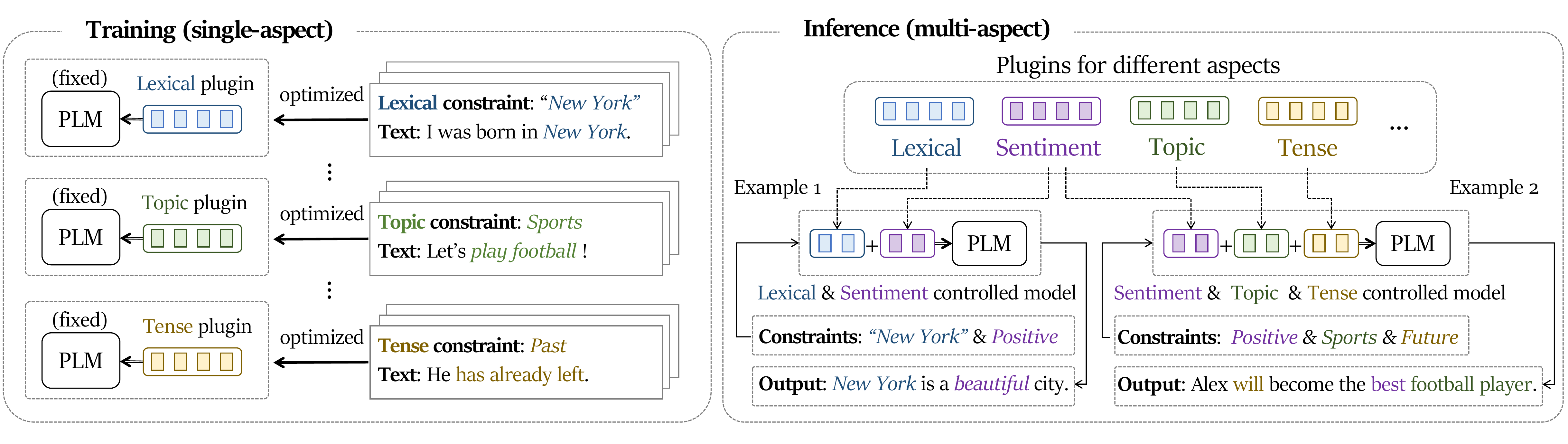}
  }
  \caption{Overview of our proposed extensible plug-and-play method for multi-aspect controllable text generation. First, plugins are trained on single-aspect labeled data separately (left). Then, arbitrary plugins can be combined by simply concatenation and plugged into the pretrained model to satisfy corresponding combinations of constraints (right). Due to the separate training of different plugins, the cost of extending a new constraint is relatively low. Besides, our approach restrains the accumulation of mutual interference, alleviating the degeneration of constraints. }
  \label{fig:extensible}
\end{figure*}

Despite their effectiveness, existing methods still suffer from low extensibility. Ideally, suppose a multi-aspect controllable text generation system has learned how to control sentiment, topic and keywords separately, it should be extensible to any combinations of the three aspects, e.g., generating a \emph{sports-themed} sentence with \emph{negative sentiment} containing \emph{keywords ``New York''} (see Figure \ref{fig:extensible}). Moreover, an extensible system should also be easily extended to control new aspects in a plug-and-play way. However, it is non-trivial for existing methods to achieve this goal. 
Specifically, the dedicated conditional generative models~\cite{keskar2019ctrl} mostly need to be trained from scratch or finetuned when facing new aspect combinations. 
The decoding-time regulation based methods~\cite{lin-riedl-2021-plug,DBLP:conf/nips/KumarMST21} intervene in the probabilities of sentences by light-weight attribute classifiers or language models during inference, which significantly impairs text fluency when multiple distributions are interpolated. 
The parameter efficient tuning based methods~\cite{qian-etal-2022-controllable,DBLP:journals/corr/abs-2210-02889} control aspects by inserting trainable prompts or prefixes into the model, referred to as plugins. By leveraging one plugin for each aspect, these methods can naturally work in a plug-and-play way, showing better potential to achieve high extensibility. 

However, existing studies show that directly combining multiple plugins  results in significantly lower controllability of the corresponding aspects than before combining (i.e., attribute degeneration)~\cite{qian-etal-2022-controllable,DBLP:journals/corr/abs-2210-02889}.
\citet{DBLP:journals/corr/abs-2210-02889} argue that \textit{mutual interference} of the plugins is the major reason for attribute degeneration, which is further justified by our theoretical and empirical analyses. Previous works alleviate the problem by introducing connectors to connect multiple plugins~\cite{DBLP:journals/corr/abs-2204-13362}, latent variables to represent the unsupervised aspects~\cite{qian-etal-2022-controllable}, or objectives to narrow the discrepancy of aspects~\cite{DBLP:journals/corr/abs-2210-02889}. However, these methods require joint training of plugins and are designed for pre-defined closed-set constraints. In consequence, their extensibility is limited.

In this paper, we propose an extensible plug-and-play method, \textsc{Prompt Gating}, for multi-aspect controllable text generation. We derive a theoretical lower bound for the interference of plugins and reveal that it accumulates with the increasing number of layers where prefixes are inserted. Based on these findings, we propose attaching trainable gates to the plugins, which normalize the interventions of plugins.
As a result, the mutual interference has been significantly reduced such that the control of arbitrary combinations of aspects can be realized by simply concatenating the corresponding plugins. Thus, our method is both extensible and plug-and-play. Moreover, we represent the constraints of the aspects in textual form, which makes our method applicable not only to categorical aspects (e.g., sentiment) but also to free-form aspects (e.g., lexical constraint).

Our contributions are three-fold:
\begin{itemize} 
    \item We propose an extensible plug-and-play method, \textsc{Prompt Gating}, for multi-aspect controllable text generation, which is able to control training-time unseen aspect combinations by simply concatenating plugins.
    \item We provide a theoretical lower bound along with empirical analyses for the mutual interference problem, which we believe will facilitate future research.
    \item Experiments show that our approach has lower mutual interference, leading to superiority over strong baselines on text quality, constraint accuracy, and extensibility.
\end{itemize}

\section{Background}

In this section, we illustrate the widely-used prefix-tuning-based method~\cite{qian-etal-2022-controllable, DBLP:journals/corr/abs-2210-02889,DBLP:journals/corr/abs-2204-13362} for multi-aspect controllable text generation. Generally, prefix-tuning~\cite{li-liang-2021-prefix} prepends light-weight continuous vectors to the multi-head attention sublayer of each Transformer layer~\cite{vaswani2017attention}:
\begin{alignat}{1}
\mathbf{H} = \mathrm{Att}\Big(\mathbf{Q},  \big[\mathbf{P}^K;\mathbf{K}\big],\big[\mathbf{P}^V;\mathbf{V}\big]\Big),
\end{alignat}
where $\mathrm{Att}(\cdot)$ is the attention function, $\mathbf{Q}$ are queries of inputs, $\mathbf{K}$ and $\mathbf{V}$ are activations of inputs, $\mathbf{P}^K$ and $\mathbf{P}^V$ are trainable prefixes, $[\cdot;\cdot]$ denotes the concatenation operation, $\mathbf{H}$ is the output of the attention sublayer. We use $\bm{\phi}$ to denote the set of prefixes in all Transformer layers.

Specifically, for multi-aspect controllable text generation, we assume that there are $N$ aspects of constraints. Due to the lack of multi-aspect labeled data, each set of prefixes, which usually represents a specific constraint (e.g., ``positive'' for the sentiment aspect), is trained on the corresponding single-aspect labeled data:
\begin{alignat}{1}
\hat{\bm{\phi}}_i &= \mathop{\mathrm{argmax}}_{\bm{\phi}_i} {\Big\{ P(\mathbf{y}|\mathbf{x}; \bm{\theta}, \bm{\phi}_i) \Big\}}, 1 \leq i \leq N, \label{eqn:train}
\end{alignat}
where $\bm{\theta}$ are the fixed parameters of the pretrained model, $\mathbf{y}$ is the controlled target sentence, $\mathbf{x}$ is the input sentence\footnote{Note that $\mathbf{x}$ is the source sentence or context for tasks like machine translation~\cite{Bahdanau:15} or summarization~\cite{nallapati-etal-2016-abstractive} and can be eliminated when it is not present.},
$P(\mathbf{y}|\mathbf{x}; \bm{\theta}, \bm{\phi}_i)$ is the conditional probability of $\mathbf{y}$, and $\hat{\bm{\phi}}_i$ are the learned parameters of prefixes for the $i$-th aspect.

During inference, for a combination of multiple aspects, corresponding prefixes are selected and synthesized by either concatenating~\cite{qian-etal-2022-controllable,DBLP:journals/corr/abs-2204-13362} or finding their intersection~\cite{DBLP:journals/corr/abs-2210-02889}, and then the generation is conditioned on the synthesis. Without loss of generality, we take two aspects as an example. The conditioned probability can be represented as
\begin{alignat}{1}
P\Big(\hat{\mathbf{y}}|\mathbf{x}; \bm{\theta}, \mathrm{syn}(\hat{\bm{\phi}}_1, \hat{\bm{\phi}}_2)\Big), \label{eqn:inference}
\end{alignat}
where $\mathrm{syn}(\cdot)$ is a synthesize function, $\hat{\mathbf{y}}$ is the candidate sentence, $\hat{\bm{\phi}}_1$ and $\hat{\bm{\phi}}_2$ are two sets of prefixes corresponding to two aspects (e.g., ``positive'' for sentiment and ``sports'' for topic), respectively.

Although existing methods alleviate the mutual interference of prefixes by joint training~\cite{qian-etal-2022-controllable, DBLP:journals/corr/abs-2210-02889,DBLP:journals/corr/abs-2204-13362}, they are based on pre-defined closed-set constraints, which increases the overhead of extending a new constraint and thus limits extensibility. Thus, to maintain high extensibility, reducing mutual interference without joint training still remains a challenge.

\section{Analyses on Mutual Interference} \label{sec:analysis}

To alleviate mutual interference while maintaining extensibility, we conduct theoretical and empirical analyses. First, we provide a definition of mutual interference as follows.

\paragraph{Definition.} 
Mutual interference (MI) is the interference between multiple plugins which are trained separately during training but are combined to guide the pretrained model simultaneously during inference (i.e., in the zero-shot setting). However, the exact interference is hard to analyze because of the complexity of deep neural networks. Intuitively, suppose multiple plugins are optimized simultaneously during training, which requires multi-aspect labeled data, their interference will be minimized because they have learned to work cooperatively under supervision (i.e., in the supervised setting). Therefore, we use the differences between the hidden states of the supervised and zero-shot settings to approximate the mutual interference of multiple plugins. Specifically, let $\hat{\bm{\phi}}_i$ and $\tilde{\bm{\phi}}_i$ be the parameters of plugins learned from the single- and multi-aspect labeled data, respectively. Taking two-aspect controlling as an example, the output of a Transformer layer is given by $\mathbf{H}(\mathbf{x}, \bm{\phi}_1, \bm{\phi}_2)$, where $\mathbf{x}$ is the layer input, then mutual interference can be defined as 
\begin{alignat}{1}
\mathrm{MI} \approx \Big\Vert\mathbf{H}(\mathbf{x}, \hat{\bm{\phi}}_1, \hat{\bm{\phi}}_2)-\mathbf{H}(\mathbf{x}, \tilde{\bm{\phi}}_1, \tilde{\bm{\phi}}_2)\Big\Vert. \label{eqn:definition}
\end{alignat}

\begin{figure}
  \centering
  \scalebox{0.4}{
  \includegraphics[]{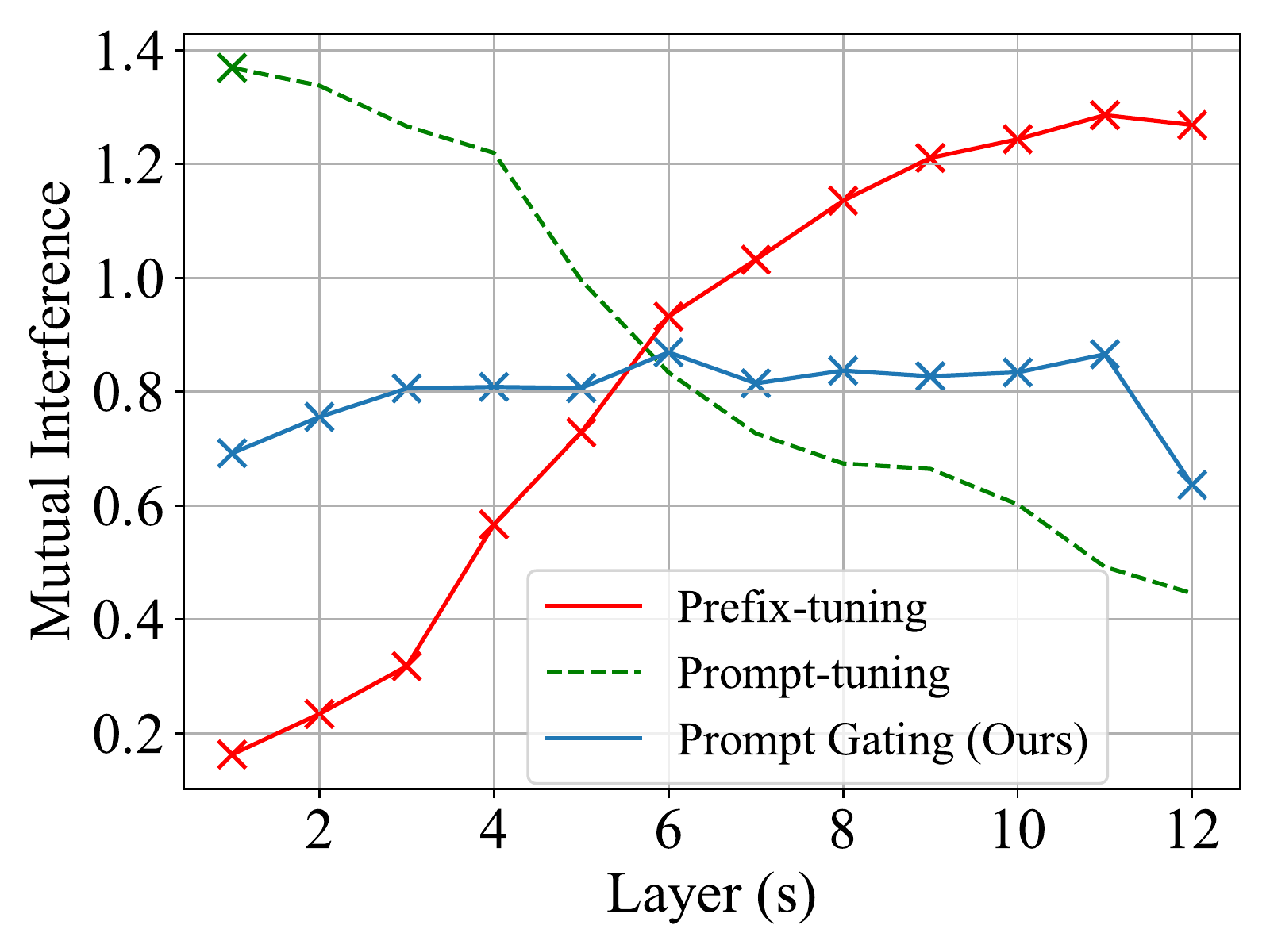}
  }
  \caption{
  The variations of mutual interference with the number of Transformer layers. Note that ``$\times$'' represents the insertion of continuous vectors. Prompt-tuning only inserts vectors into the model after the embedding layer, while the other two methods insert vectors into each Transformer layer. Our approach (Prompt Gating) restrains the growth of mutual interference while inserting sufficient trainable parameters. 
  }
  \label{fig:interference}
\end{figure}

\paragraph{Empirical Analysis.}
Then, as mutual interference has been defined as the norm of gap between hidden states in the zero-shot and supervised settings, we can empirically estimate it on the authentic dataset. By calculating the averaged norm on the Yelp dataset~\cite{DBLP:conf/iclr/LampleSSDRB19}, we plot the variations of mutual interference with the number of Transformer layers for Prompt-tuning~\cite{lester-etal-2021-power} and Prefix-tuning~\cite{li-liang-2021-prefix} in Figure \ref{fig:interference}. 
We can find that the interference accumulates with insertions of trainable parameters. Moreover, the magnitude of mutual interference at the last Transformer layer (i.e., the $12$-th layer in Figure \ref{fig:interference}) is consistent with the performance gap, which is the difference between the fulfillment of constraints in single- and multi-aspect settings (see Table \ref{tab:ctg}). Meanwhile, too few trainable parameters cannot guide the pretrained model effectively. In summary, the key point for remaining effective in the zero-shot setting is \textit{restraining the growth of mutual interference} (for a lower performance gap) while \textit{providing sufficient trainable parameters} (for better supervised performance).

\paragraph{Theoretical Analysis.}
Next, to find a way to alleviate mutual interference, we conducted a theoretical analysis.\footnote{Please refer to Appendix \ref{app:theory} for more detail.} As a result, we found that the mutual interference, which is caused by the interactions in attention sublayers, has a theoretical lower bound\footnote{For brevity, we show the lower bound in one head of attention (obtained by simplifying Eq.~(\ref{eqn:mi_singlehead})), and a similar conclusion can be obtained on the multi-head attention (Eq.~(\ref{eqn:mi_multihead})).}:
\begin{alignat}{1}
\mathrm{MI} > \alpha\Vert\Delta\mathbf{h}_1(\mathbf{x}, \hat{\bm{\phi}}_1)\Vert+\beta\Vert\Delta\mathbf{h}_2(\mathbf{x}, \hat{\bm{\phi}}_2)\Vert, \label{eqn:lower_bound}
\end{alignat}
where $0 < \alpha, \beta < 1$, and $\Vert\Delta\mathbf{h}_i(\mathbf{x}, \hat{\bm{\phi}}_i)\Vert$ is a norm that is positively related to the magnitude of $\hat{\bm{\phi}}_i$. Moreover, the lower bound might accumulate with Transformer layers like in Figure \ref{fig:interference}.
Intuitively, applying normalization (e.g., gates) to the parameters of the $i$-th plugin to reduce its magnitude will decrease the lower bound of mutual interference.

\section{\textsc{Prompt Gating}} \label{sec:gating}

We propose a novel approach that attaches trainable gates to the plugins, which alleviates the mutual interference of multiple plugins and makes the model highly extensible. Figure \ref{fig:method} shows the architecture of our approach. We first provide intuition in \S\ref{sec:intuition}, then define our approach formally in \S\ref{sec:method}.

\subsection{Intuition} \label{sec:intuition}

\begin{figure}[!t]
  \centering
  \resizebox{\linewidth}{!}{
  \includegraphics[]{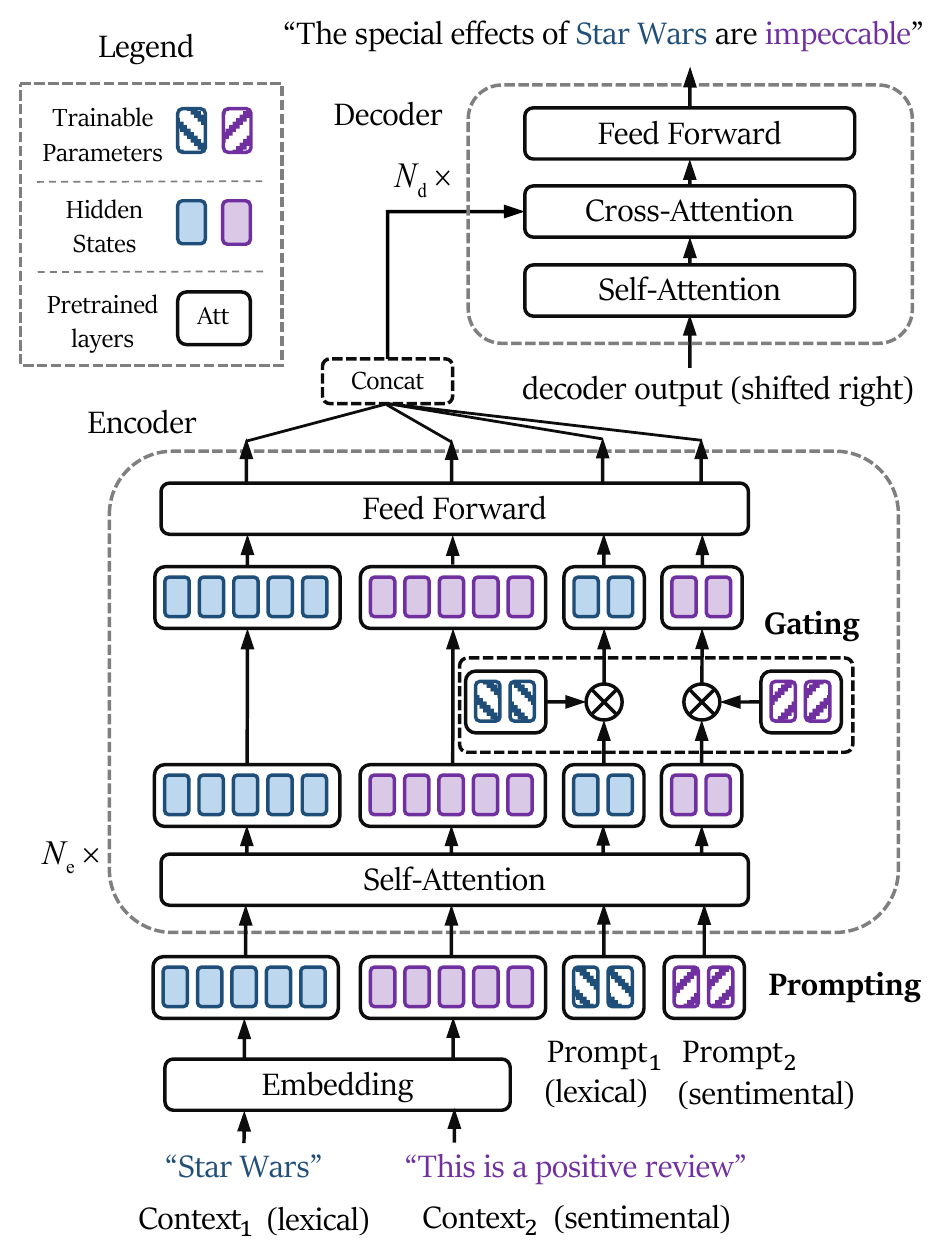}
  }
  \caption{The architecture of our approach. It shows the case of inference stage of double-aspect controlled text generation. Blue and purple represent lexical and sentimental constraints respectively. Continuous prompts and contextual contexts are fed into the model and  trainable gates are applied to steer the pretrained model as well as alleviate the mutual interference of plugins.} 
  \label{fig:method}
\end{figure}

Although prefix-tuning provides sufficient interventions and avoids long-range dependencies by inserting continuous vectors into each attention sublayer, it suffers from the accumulation of mutual interference of these plugins (see \S\ref{sec:analysis}). On the one hand, the vectors are inserted into the attention sublayer, where they interact with each other, which directly enhances mutual interference. On the other hand, the vectors are not normalized, which leads to a large lower bound of mutual interference (Eq.~(\ref{eqn:lower_bound})). Intuitively, injecting the vectors in a position-wise manner will avoid direct interaction between them. Moreover, normalizing the vectors can limit the magnitude of the lower bound, which might decrease mutual interference. 
Therefore, we first propose attaching vectors outside the attention sublayer, which can be realized by appending trainable vectors to the output of the embedding layer and adding trainable vectors to the hidden states in each Transformer layer (see Figure \ref{fig:method}). Then, trainable gates are applied to these hidden states to alleviate mutual interference further. In this way, we expect our approach to restrain the growth of mutual interference while providing sufficient interventions.

\subsection{Method} \label{sec:method}

\paragraph{Prompting.}
We present our model in the order of forward propagation. To change how the trainable parameters are injected into the model, we first follow prompt-tuning~\cite{lester-etal-2021-power} to append trainable prompts to the output of the embedding layer. Moreover, to make our model applicable not only to categorical aspects (e.g., sentiment) but also to free-form aspects (e.g., lexical constraint), we present the constraints of aspects in textual form and feed them to the model. When two aspects of constraints are required during inference, the model input is given by 
\begin{alignat}{1}
\mathbf{H}^{(0)} = \Big[\mathrm{E}(\mathbf{x});\mathrm{E}(\mathbf{c}_1);\mathrm{E}(\mathbf{c}_2);\mathbf{P}^{(0)}_1;\mathbf{P}^{(0)}_2\Big], \label{eqn:input}
\end{alignat}
where $\mathrm{E}(\cdot)$ is the embedding function, and $\mathbf{x}$ is the source sentence for sequence-to-sequence generation like machine translation and can be eliminated for text generation. $\mathbf{c}_1$ and $\mathbf{c}_2$ are textual form of constraints (e.g., ``This is a positive review'' for positive review generation, and ``New York'' for lexical constraint). $\mathbf{P}^{(0)}_1, \mathbf{P}^{(0)}_2 \in \mathbb{R}^{p \times d}$ are continuous prompts, where the right superscript $^{(j)}$ represents the $j$-th layer, $p$ is the number of continuous vectors, and $d$ is the dimension of hidden states. To avoid the discrepancy between training and inference in position, each textual sequence has its own position indexes starting from 1 and its own segment embedding~\cite{devlin2019bert}. Note that only one textual constraint and one set of trainable parameters are injected during training.

\paragraph{Gating.}
Then, the model input $\mathbf{H}^{(0)}$ is fed to the encoder, where trainable gates are attached to the hidden states in a position-wise manner, which alleviates mutual interference as well as steers the model.
Formally, $\mathbf{A}^{(j)} =\mathrm{Self\text{-}Att}(\mathbf{H}^{(j-1)})$ is the output of the $j$-th attention sublayer, and it is normalized by the gates:

\begin{alignat}{1}
\Tilde{\mathbf{A}}^{(j)} = \Big[&\mathbf{A}^{(j)}_X;\sigma\big(\mathbf{G}^{(j)}_1\big)\odot \big(\mathbf{A}^{(j)}_{P_1} +\mathbf{P}^{(j)}_1\big); \nonumber \\
& \sigma\big(\mathbf{G}^{(j)}_2\big)\odot \big(\mathbf{A}^{(j)}_{P_2} +\mathbf{P}^{(j)}_2\big) \Big], \label{eqn:gate}
\end{alignat}
where $\mathbf{A}^{(j)}_X \in \mathbb{R}^{(|\mathbf{x}|+|\mathbf{c}_1|+|\mathbf{c}_2|) \times d}$ and $\mathbf{A}^{(j)}_{P_i} \in \mathbb{R}^{p \times d}$ are hidden states split from $\mathbf{A}^{(j)}$, $\mathbf{P}_i^{(j)} \in \mathbb{R}^{p \times d}$ are trainable vectors that add to the hidden states, $\sigma$ is the $\mathrm{sigmoid}(\cdot)$ function and $\mathbf{G}_i^{(j)} \in \mathbb{R}^{p \times d}$ are trainable vectors. $\odot$ denotes the Hadamard product and the normalized vectors $\sigma(\mathbf{G}_i^{(j)})$ serve as {\bf gates} to selectively rescale the output of the attention sublayer in a position-wise manner and  $\Tilde{\mathbf{A}}^{(j)} \in \mathbb{R}^{(|\mathbf{x}|+|\mathbf{c}_1|+|\mathbf{c}_2|+2p) \times d}$ is the result of the normalization. Next, the normalized output is fed to the feed-forward sublayer: $\mathbf{H}^{(j)} =\mathrm{FFN}(\Tilde{\mathbf{A}}^{(j)})$. Finally, the output of the last encoder layer is fed to a standard Transformer decoder to guide the generation.

\paragraph{Training \& Inference.} As shown in Figure \ref{fig:extensible}, during training, each plugin (including prompts and gates) for a single aspect of constraints is attached to the pretrained generative model and optimized by corresponding single-aspect labeled data separately (refer to Eq.~(\ref{eqn:train})). In contrast, during inference, the control of arbitrary combinations of aspects can be realized by simply concatenating the corresponding plugins (refer to Eq.~(\ref{eqn:inference})).

Moreover, our approach treats the training and inference processes for pre-existing and newly-introduced constraints identically. The total training cost of $N$ pre-existing aspects and $M$ newly-added aspects is $O((N+M)C)$, where $C$ denotes the cost of training on a single aspect. In this way, the cost of introducing new constraints is relatively low.

\section{Experiments}

We conducted experiments on two representative tasks in natural language generation, which are text generation and machine translation.

\subsection{Multi-Aspect Controllable Text Generation}
\paragraph{Dataset.}\label{sec:ctg_data}
Following previous work~\cite{DBLP:journals/corr/abs-2204-13362}, we adopted the widely-used Yelp dataset~\cite{DBLP:conf/iclr/LampleSSDRB19}, which contains restaurant reviews with the sentiment (positive and negative) and the topic (American, Mexican, and Asian) labels. To evaluate the extensibility of methods, we added two additional aspects of constraints: keywords~\cite{DBLP:conf/emnlp/He21} and tense (past and present)~\cite{ficler-goldberg-2017-controlling}, where their labels are automatically extracted from the reviews. Due to the page limit, please refer to Appendix \ref{app:model_setup} for more details about the experimental setup.

\paragraph{Evaluation.}\label{sec:ctg_eval}
Following previous work, we adopted automatic and human evaluations for constraint accuracy and text quality~\cite{lyu-etal-2021-styleptb, dathathri2019plug, DBLP:journals/corr/abs-2210-02889}. Specifically, we finetuned two RoBERTa-based~\cite{liu2019roberta} classifiers for the evaluations of sentiment and topic. The tense accuracy was evaluated by the same tool adopted in the training set, and we used word-level Copy Success Rate (CSR)~\cite{chen2021lexical} to evaluate the lexical constraint. In addition, we used the perplexity (PPL) given by GPT-2$_\mathrm{medium}$~\cite{radford2019language} and averaged distinctness~\cite{li-etal-2016-diversity} to evaluate the fluency and diversity of the generated text, respectively. For human evaluation, each sentence received a score of 1 to 5 on sentiment and topic relevance as well as fluency given by three evaluators. The final scores are averaged over three ratings.

\begin{table*}[!t]
    \centering
    \small
    \begin{center}
        \begin{tabular}{ll|c|ccc|c}
            \toprule
            {\bf Category} &
             {\bf Method} &
             \bf Dist.$\uparrow$ &
             \bf Sent.$\uparrow$ &
             \bf Topic$\uparrow$  &
             \bf Average$\uparrow$ &
             \bf PPL$\downarrow$ 
            \\ \midrule \midrule

            \em DTR &
             \textsc{GeDi} &
             {0.75\ \scriptsize{\ (0.00)}} &
             {\bf 99.47}\ \scriptsize{\bf\ (+0.13)} &
             {51.36\ \scriptsize{(-45.98)}}  &
             {75.41\ \scriptsize{(-22.92)}} &
             {616.92\ \scriptsize{(+253.23)}} \\
            
            \midrule
            
            \em PET w/ JT &
             \textsc{Dist. Lens} &
             {0.26\ \scriptsize{(-0.10)}} &
             {77.47\ \scriptsize{(-17.17)}} &
             {66.98\ \scriptsize{(-14.95)}}&
             {72.22\ \scriptsize{(-16.06)}}  &
             {\ \ 52.59\ \scriptsize{\ \ (+19.73)}} \\

            \midrule
            
            \multirow{4}{*}{\em PET w/o JT} &
             \textsc{Prompt-tuning} &
             0.42\ \scriptsize{(-0.06)} &
             48.29\ \scriptsize{\ \ (-5.84)} &
             48.11\ \scriptsize{\ \ (-8.82)} &
             48.20\ \scriptsize{\bf\ \ (-7.33)} &
             \ \ 40.89\ \scriptsize{\bf\ \ \ \ \ (-6.83)} \\
            
            &
             \textsc{Prefix-tuning} &
             0.31\ \scriptsize{(-0.10)} &
             47.53\ \scriptsize{(-37.27)} &
             69.11\ \scriptsize{\ \ (-9.38)} &
             58.32\ \scriptsize{(-23.32)} &
             147.47\ \scriptsize{(+125.17)} \\

            &
             \textsc{Tailor} &
             {0.39}\ \scriptsize{(-0.04)} &
             {80.68}\ \scriptsize{\ \ (-8.12)} &
             {68.72}\ \scriptsize{\ \ (-9.94)} &
             {74.70}\ \scriptsize{\ \ (-9.03)} &
             \ \ {40.29}\ \ \ \ \scriptsize{(+8.52)} \\

             \cmidrule(lr){2-7}
            &
             \textsc{Prompt Gating} (\textit{Ours}) &
             {0.42\ \scriptsize{\ (0.00)}} &
             {84.80\ \scriptsize{(-10.93)}} &
             {\bf75.02\ \scriptsize{\ \ (-8.00)}} &
             {\bf 79.91}\ \scriptsize{\ \ (-9.47)} &
             \ \ {\bf21.77}\ \ \ \ \scriptsize{(+0.14)} \\
            
            \bottomrule
        \end{tabular}
        \caption{\label{tab:ctg} Automatic evaluation on double-aspect controllable text generation. ``{\em DTR}'' and ``{\em PET}'' denote decoding-time regulation and parameter efficient tuning methods, respectively.  ``{\em w/ JT}'' and ``{\em w/o JT}'' denote methods with and without joint training, respectively. There are two aspects: ``Sent.'' (sentiment) and ``Topic''. ``Average'' denotes the averaged scores over sentiment and topic accuracies. ``PPL'' and ``Dist.'' denote perplexity and averaged distinctness, respectively. The scores in brackets indicate the performance gap between double- and single-aspect settings.}

    \end{center}
\end{table*}

\begin{table*}[!t]
    \centering
    \small
    \begin{center}
        \resizebox{\linewidth}{!}{
        \begin{tabular}{cl|l|lllcc|c}
            \toprule
            \bf \#~Aspects & 
            \bf Method & 
            \bf \ \  PPL $\downarrow$ & 
            \bf  Sent. $\uparrow$ & 
            \bf  Topic $\uparrow$& 
            \bf  Tense $\uparrow$& 
            \bf Lex. $\uparrow$ & 
            \bf Ave.$\uparrow$ & 
            \bf $\Delta$Time \\
            \midrule \midrule

            \multirow{3}{*}{3} 
            & 
            \textsc{Prefix-tuning}  & 
            154.69  & 
            44.91 & 
            54.38 & 
            24.49 & 
            / & 
            41.26 & 
            \ \ +6.42 h \\
            
            & 
            \textsc{Dist. Lens} & 
            \ \ 63.13 & 
            65.31 & 
            55.84 & 
            54.25 & 
            / & 
            58.47 & 
            +30.13 h \\

            & 
            \textsc{Prompt Gating} (\textit{Ours}) & 
            \ \ {\bf21.87}  & 
            {\bf76.93} & 
            {\bf62.73} & 
            {\bf62.24} & 
            / & 
            \bf 67.30 & 
            \bf \ \ +6.30 h \\
            
            \midrule

            \multirow{2}{*}{4} & 
            \textsc{Prefix-tuning}  & 
            159.80 \scriptsize{\ \ (+8.35)} & 
            37.33 \scriptsize{\ \ (-7.58)} & 
            32.51 \scriptsize{(-21.87)} & 
            18.82 \scriptsize{(-5.68)} & 
            29.55 & 
            15.47 & 
            \ \ +2.77 h \\
            
            & 
            \textsc{Prompt Gating} (\textit{Ours}) & 
            \ \ \bf20.90 \scriptsize{\ \ \ (-0.96)} & 
            \bf75.32 \scriptsize{\ \ (-1.61)} & 
            \bf62.52 \scriptsize{\ \ (-0.21)} & 
            \bf60.05 \scriptsize{(-2.20)} & 
            \bf54.50 & 
            \bf 63.10 & 
            \bf \ \ +2.01 h \\

            \bottomrule
        \end{tabular}
        }
        \caption{\label{tab:extensible} Automatic evaluation on triple- and quadruple-aspect controllable text generation where the models are extended from double-aspect setting. ``\#~Aspects'' denotes the number of aspects ($N$). ``Ave.'' denotes the averaged accuracies over $N$ aspects. ``Sent.'', ``Topic'', and ``Tense'' denote accuracies for sentimental, topical, and temporal constraints, respectively. ``Lex.'' denotes CSR for  lexical constraint. ``$\Delta$Time'' denotes the training time extending from $(N-1)$-aspect setting to $N$-aspect setting. The scores in brackets indicate the performance gap between quadruple- and triple-aspect settings. Note that the methods specialized for attribute-based controlling like \textsc{Dist. Lens} can not process free-form constraints like lexical constraint.}
    \end{center}
\end{table*}

\begin{table}[!t]
    \centering
    \small
    \begin{center}
        \resizebox{\linewidth}{!}{
        \begin{tabular}{l|cc|c}
            \toprule
            \bf Method & \bf Sent. $\uparrow$ & \bf Topic $\uparrow$& \bf Fluency $\uparrow$ \\
            \midrule \midrule
            
            \textsc{GeDi} & 1.67 & 2.72 & 3.12 \\

            \textsc{Dist. Lens} & 3.71 & 3.20 & 3.72 \\
            
            \textsc{Prompt Gating} (\textit{Ours}) & \bf 4.44 & \bf 4.23 & \bf 4.19\\
            
            \bottomrule
        \end{tabular}
        }
        \caption{\label{tab:human} Human evaluation on double-aspect controllable text generation. Sentences are rated 1 to 5 each for sentimental and topical relevance and fluency.}
    \end{center}
\end{table}

\paragraph{Baselines.}\label{sec:ctg_baseline}
We compared our approach with several representative methods for multi-aspect controllable text generation:
\begin{itemize}

\item \textsc{GeDi}~\cite{krause-etal-2021-gedi-generative}: a decoding-time regulation method that uses light-weight conditional generative discriminator to guide pretrained models. The distributions given by multiple discriminators are normalized for controlling multiple aspects of target sentences. 

\item  \textsc{Dist. Lens}~\cite{DBLP:journals/corr/abs-2210-02889}: a prefix-tuning-based method that introduces an autoencoder and additional objectives to map several constraints of attributes to one latent space (i.e., joint training of prefixes). It finds the intersection of prefixes of multiple constraints during inference.

\item  \textsc{Prompt-tuning}~\cite{lester-etal-2021-power}: a parameter efficient method that appends continuous prompts to the model input. Multiple prompts are trained separately and are simply concatenated during inference. 

\item  \textsc{Prefix-tuning}~\cite{li-liang-2021-prefix}: a parameter efficient method that appends continuous prefixes to the activations at attention sublayers. Multiple prefixes are trained separately and are simply concatenated during inference. 

\item \textsc{Tailor}~\cite{DBLP:journals/corr/abs-2204-13362}: a prompt-tuning-based method that further modifies the attention mask and position indexes during inference to narrow the gap between training and inference.
\end{itemize}

\begin{table*}[!t]
    \centering
    \small
    \begin{center}
        \begin{tabular}{l|ccc|c}
            \toprule
            \bf Method &
             \bf Lex. $\uparrow$ &
             \bf Tense$\uparrow$ &
             \bf Average$\uparrow$ &
             \bf BLEU$\uparrow$ \\
            \midrule \midrule
            
            \textsc{Prefix-tuning} &
             \ \ 7.51 \scriptsize{(-77.77)} &
             43.46 \scriptsize{(-39.83)} &
             25.48 \scriptsize{(-58.80)} &
             \ \ 0.4 \scriptsize{(-36.3)} \\

            \textsc{Parallel Adapter} &
             48.44 \scriptsize{(-43.38)} &
             67.87 \scriptsize{(-15.68)} &
             58.15 \scriptsize{(-29.53)} &
             21.8 \scriptsize{(-15.5)} \\

            \textsc{LoRA} &
             50.79 \scriptsize{(-37.15)} &
             74.16 \scriptsize{(-10.16)} &
             62.47 \scriptsize{(-23.65)} &
             25.0 \scriptsize{(-11.2)} \\
            
            \textsc{Prompt-tuning} &
             64.64 \scriptsize{(-10.29)} &
             81.12 \scriptsize{\ \ (-0.07)} &
             72.88 \scriptsize{\ \ (-5.18)} &
             34.2 \scriptsize{\ \ (-1.2)} \\

            \midrule
            \textsc{Prompt Gating} (\textit{Ours})&
             {\bf85.29} \scriptsize{\ \ (-4.61)} &
             {\bf85.75} \scriptsize{\ (+1.76)} &
             {\bf85.52} \scriptsize{\ \ (-1.42)} &
             {\bf36.8} \scriptsize{\ \ (-0.3)}  \\
            
            \bottomrule
        \end{tabular}
        \caption{\label{tab:mt} Results on controllable machine translation. The experiments are conducted on the WMT14 German$\rightarrow$English benchmark. There are three aspects of constraints: lexical constraint, tense, and external knowledge (French synonymous sentences). ``Lex.'' and ``Tense'' denote CSR and accuracies for lexical and temporal constraint, respectively. ``Average'' denotes the averaged accuracies over them. The scores in brackets indicate the performance gap between double- and single-aspect settings.
        }
    \end{center}
\end{table*}

\paragraph{Results.}
Table \ref{tab:ctg} shows the automatic evaluation on double-aspect controllable text generation. We demonstrate the averaged accuracies to represent the overall performance on satisfying multiple aspects of constraints. Furthermore, we provide the performance gap between double- and single-aspect settings to represent the ability to combine multiple plugins in a zero-shot manner. Although \textsc{GeDi} achieves the highest scores on sentiment accuracy and distinctness, its perplexity explodes, and its tense accuracy is significantly decreased, which can be attributed to the interpolation of multiple discriminators. As \textsc{Prompt-Tuning} does not have sufficient trainable parameters, it performs poorly on constraint accuracies. However, it has a relatively minor performance gap due to only inserting vectors once. \textsc{Prefix-tuning} suffers from severe mutual interference because of the insertions in all Transformer layers, leading to poor performance on either constraint accuracies or perplexity. Compared with \textsc{Prefix-tuning}, \textsc{Dist. Lens} has better constraint accuracies and lower performance gaps because of the joint training of prefixes. We found that \textsc{Dist. Lens} is sensitive to constraint distributions in the training set because it attempts to find the intersection of multiple constraints. Our approach (\textsc{Prompt Gating}) achieves the highest constraint accuracies, lowest perplexity and a relatively small performance gap while our plugins are trained separately.

Table \ref{tab:extensible} shows the extensibility of the methods. When extended from double-aspect to triple-aspect, \textsc{Dist. Lens} has to be retrained because of its joint training strategy. In contrast, our approach and \textsc{Prefix-tuning} only need to train one plugin, then combine plugins and plug them into the pretrained model. Unfortunately, when extended from triple-aspect to quadruple-aspect, as plugins of \textsc{Prefix-tuning} badly interfere with each other, its ability to control multiple aspects significantly degenerates. However, our approach has a slight performance gap with a relatively small training cost, revealing its high extensibility.

The human evaluation results are illustrated in Table \ref{tab:human} with an inter-annotator agreement of 0.31 (Fleiss' $\kappa$). Experiments indicate that our approach significantly outperforms both baselines with $p<0.01$ on all three aspects, determined by paired bootstrap and t-test using a popular open-source tool~\cite{dror-etal-2018-hitchhikers}\footnote{\url{https://github.com/rtmdrr/testSignificanceNLP}}. Unlike automatic evaluations, \textsc{GeDi} performs the worst in sentiment relevance. It can probably be attributed to the fact that \textsc{GeDi} often generates ambivalent-sentiment and non-fluent sentences, and human annotators tend to give low ratings to them. The other results are in line with automatically evaluated results.

\subsection{Multi-Aspect Controllable Machine Translation}
\paragraph{Dataset.}
To thoroughly compare our approach with baselines, we also adopted a sequence-to-sequence generation task (i.e., machine translation~\cite{Bahdanau:15}). Experiments are conducted on the WMT14 German $\rightarrow$ English benchmark. We adopted three aspects of constraints in machine translation, and the labels are all automatically obtained from target sentences. We use keywords~\cite{post-vilar-2018-fast} and tense~\cite{ficler-goldberg-2017-controlling} like in the text generation task to control translations. Specifically, we adopt French sentences with the same meaning as the German sources, which can be seen as an external knowledge to improve translation quality~\cite{zoph2016multi}, as the third constraint. 

\paragraph{Evaluation.} We adopted \textsc{SacreBLEU}\footnote{The signature is ``BLEU+case.mixed+numrefs.1+smooth .exp+tok.13a+version.2.0.0''.} \cite{post2018call} to calculate BLEU scores~\cite{papineni-etal-2002-bleu} to evaluate the translation quality. Similar to text generation (\S\ref{sec:ctg_eval}), we used CSR~\cite{chen2021lexical} and tense accuracy to evaluate lexical and temporal constraints, respectively.

\paragraph{Baselines.} Besides \textsc{Prompt-tuning}~\cite{lester-etal-2021-power} and \textsc{Prefix-tuning}~\cite{li-liang-2021-prefix} (\S\ref{sec:ctg_baseline}), we adopted another two representative parameter efficient methods as baselines:
\begin{itemize}

\item  \textsc{LoRA}~\cite{hu2021lora}: a method that adds trainable rank decomposition matrices into attention sublayers.

\item  \textsc{Parallel Adapter}~\cite{houlsby2019parameter}: a method that parallelly inserts feed-forward sublayers between pre-trained sublayers. 

\end{itemize}
Similar to \textsc{Prefix-tuning}, for both \textsc{LoRA} and \textsc{Parallel Adapter}, each plugin is trained separately, and multiple plugins are simply concatenated for multi-aspect setting during inference.

\paragraph{Results.}
Table \ref{tab:mt} shows the results on controllable machine translation. Unlike text generation, constraints in machine translation do not merely contain attribute-based constraints. Therefore, methods specially designed for attribute-based constraints cannot be applied to this task. Surprisingly, \textsc{Prompt-tuning} achieves the highest constraint accuracies and translation quality among baselines because it largely retains the capabilities of plugins to satisfy constraints. \textsc{Prefix-tuning} faces the severe degeneration of both accuracies of constraints and BLEU scores, which might be attributed to the more complicated model structure in machine translation than text generation. Our approach outperforms all baselines in machine translation, and the consistent superiorities on both tasks show its generalizability.

\begin{table}[!t]
    \centering
    \small
    \begin{center}
        \resizebox{\linewidth}{!}{
        \begin{tabular}{l|cc|c}
            \toprule
            \bf Method & \bf Sent. $\uparrow$ & \bf Topic $\uparrow$& \bf PPL $\downarrow$ \\
            \midrule \midrule
            
            \textsc{Prompt Gating} (\textit{Ours}) & \bf 84.80 & \bf 75.02 & \bf 21.77 \\
            \midrule
            - Textual context for attribute & 83.60 & 71.89 & 22.00 \\
            
            - Normalization of gates & 76.53 & 61.02 & 27.55 \\
            
            Move the gates behind FFN & 56.71 & 32.49 & 36.74 \\

            \bottomrule
        \end{tabular}
        }
        \caption{\label{tab:ablation} Ablation study and comparison with the variant of our approach. ``- Textual context for attribute'' denotes ablating textual contexts of attribute-based constraints (see Eq.~(\ref{eqn:input})). ``- Normalization of gates'' denotes ablating the $\mathrm{sigmoid}(\cdot)$ function which normalizes the gates (see Eq.~(\ref{eqn:gate})). ``Move the gates behind FFN'' denotes changes where trainable gates apply.}
    \end{center}
\end{table}

\subsection{Analysis}
\paragraph{Mutual Interference.} \label{sec:analysis_mi}
Similar to empirical analysis on mutual interference for \textsc{Prefix-tuning} and \textsc{Prompt-tuning} (see \S\ref{sec:analysis}), we also plotted the variation of the mutual interference with the number of injections of our approach in Figure \ref{fig:interference}. With the gates to selectively rescale the interventions of plugins, the growth of interference is restrained.

\paragraph{Ablation Study.}
Table \ref{tab:ablation} shows the ablation study and comparison with the variant of our approach. According to the performance gaps corresponding to the changes, we can find that the textual context of constraints slightly affects the constraint accuracies, and the normalization of the trainable gates is a key point for good performance. Moreover, the trainable gates should be placed where the interactions have just happened (i.e., after attention sublayers). Please refer to Appendix \ref{sec:app_exp} and \ref{app:case_study} for more results, analyses, and cases.

\section{Related Work}

Multi-aspect controllable text generation (MCTG)~\cite{qian-etal-2022-controllable,DBLP:journals/corr/abs-2204-13362,DBLP:journals/corr/abs-2210-02889} that simultaneously satisfies multiple constraints is a challenging task for which highly extensible methods make more practical sense. Approaches to it can be roughly divided into the following three categories. 

\paragraph{Dedicated Model.}
The dedicated conditional generative models~\cite{keskar2019ctrl,dou2021gsum,huang2021transfer,chen2021lexical} can accept multiple constraints by training from scratch or full-parameter finetuning on the multi-aspect labeled data. However, the multi-aspect labeled data is hard to obtain, and the constraints that can be satisfied are already determined during training. Thus it is usually too expensive to apply dedicated models to MCTG.

\paragraph{Decoding-Time Regulation.}

Although multi-aspect controlling can be achieved by interpolating distributions of multiple discriminators~\cite{dathathri2019plug,DBLP:conf/aaai/Chen0L21a,krause-etal-2021-gedi-generative,lin-riedl-2021-plug} or optimizing towards multiple objectives~\cite{DBLP:journals/corr/abs-2202-11705,DBLP:conf/nips/KumarMST21}, they usually significantly impair text fluency because of the intervention in the decoding stage~\cite{DBLP:journals/corr/abs-2210-02889}.

\paragraph{Parameter Efficient Tuning.}
Unlike the above two branches, PET introduces plugins trained with fixed pretrained models for generating required text~\cite{li-liang-2021-prefix,lester-etal-2021-power,DBLP:conf/acl/WangTL22}. Because of its potential to achieve high extensibility in a plug-and-plug manner, our work also falls in this line. However, when multiple constraints are required, joint training of plugins is introduced to alleviate the mutual interference of plugins~\cite{DBLP:conf/iclr/ChanOPZF21,qian-etal-2022-controllable,DBLP:journals/corr/abs-2204-13362,DBLP:journals/corr/abs-2210-02889}, which hurts extensibility. Differently, our work aims at reducing mutual interference while maintaining separate training. Similar to our work, \citet{DBLP:journals/corr/abs-2204-13362} proposes preventing two prompts from interactions in attention layers by modifying attention masks. Nevertheless, their method like prompt-tuning~\cite{lester-etal-2021-power} only introduces trainable parameters to the model input, leading to insufficient trainable parameters and dissatisfied constraints. In contrast, we propose a novel PET method that attaches trainable gates to the pretrained model, alleviating mutual interference while providing sufficient interventions, leading to both desired extensibility and effectiveness.

\section{Conclusion}
In summary, we propose an extensible plug-and-play method for multi-aspect controllable text generation. By replacing trainable prefixes with trainable prompts and gates, our approach alleviates the mutual interference of multiple plugins while providing sufficient interventions. Experiments on text generation and machine translation show its superiorities over baselines on the cost of extending to new combinations of aspects, the fulfillment of constraints, and text fluency.

\section*{Limitations}
First, although our approach and existing methods for controllable text generation can improve the constraint accuracies, they are currently unable to achieve 100\% accuracies in the vast majority of aspects (e.g., sentiment or topic). This makes them not yet applicable in scenarios with requirements of 100\% control fulfillment.
Second, there is still a gap between the automatic and human evaluation of text generation, which makes there a trade-off between precision and efficiency in the evaluation of controllable text generation.
Third, although our approach reduces the mutual interference of plugins so that multiple plugins can be combined at a relatively small cost (a decrease in constraint accuracy), this cost will not be zero, which puts an upper limit on the number of plugins can be applied simultaneously. Fortunately, for controllable text generation, the number of controls applied simultaneously is generally not too large (e.g., four or five aspects).

\section*{Ethics Statement}
Since the text generation model is trained on data collected from the web and often not thoroughly cleaned, it can generate offensive or toxic text. We must state that the texts generated by our approach do not represent our opinion. To alleviate these issues, we can take detoxification and politeness as the default aspects of constraints in our multi-aspect controllable method.

\section*{Acknowledgments}
This work is supported by the National Key R\&D Program of China (2022ZD0160502), the National Natural Science Foundation of China (No. 61925601, 62276152), the National Social Science Fund of China (20\&ZD279), and a grant from the Guoqiang Institute, Tsinghua University. We thank Kaiyu Huang, Fuchao Wei, Yuanhang Zheng and all the anonymous reviewers for their valuable comments and suggestions on this work, as well as all the volunteers who participated in the human evaluation.

\bibliography{custom}

\begin{thebibliography}{41}
\expandafter\ifx\csname natexlab\endcsname\relax\def\natexlab#1{#1}\fi

\bibitem[{Bahdanau et~al.(2015)Bahdanau, Cho, and Bengio}]{Bahdanau:15}
Dzmitry Bahdanau, Kyunghyun Cho, and Yoshua Bengio. 2015.
\newblock \href {http://arxiv.org/abs/1409.0473} {Neural machine translation by
  jointly learning to align and translate}.
\newblock In \emph{Proceedings of ICLR 2015}.

\bibitem[{Chan et~al.(2021)Chan, Ong, Pung, Zhang, and
  Fu}]{DBLP:conf/iclr/ChanOPZF21}
Alvin Chan, Yew{-}Soon Ong, Bill Pung, Aston Zhang, and Jie Fu. 2021.
\newblock \href {https://openreview.net/forum?id=VD\_ozqvBy4W} {Cocon: {A}
  self-supervised approach for controlled text generation}.
\newblock In \emph{Proceedings of {ICLR} 2021}.

\bibitem[{Chen et~al.(2021)Chen, Chen, and Li}]{DBLP:conf/aaai/Chen0L21a}
Guanhua Chen, Yun Chen, and Victor O.~K. Li. 2021.
\newblock \href {https://ojs.aaai.org/index.php/AAAI/article/view/17496}
  {Lexically constrained neural machine translation with explicit alignment
  guidance}.
\newblock In \emph{Proceedings of {AAAI} 2021}.

\bibitem[{Chen et~al.(2020)Chen, Chen, Wang, and Li}]{chen2021lexical}
Guanhua Chen, Yun Chen, Yong Wang, and Victor O.~K. Li. 2020.
\newblock \href {https://doi.org/10.24963/ijcai.2020/496}
  {Lexical-constraint-aware neural machine translation via data augmentation}.
\newblock In \emph{Proceedings of IJCAI 2020}.

\bibitem[{Dathathri et~al.(2019)Dathathri, Madotto, Lan, Hung, Frank, Molino,
  Yosinski, and Liu}]{dathathri2019plug}
Sumanth Dathathri, Andrea Madotto, Janice Lan, Jane Hung, Eric Frank, Piero
  Molino, Jason Yosinski, and Rosanne Liu. 2019.
\newblock \href {https://arxiv.org/abs/1912.02164} {Plug and play language
  models: A simple approach to controlled text generation}.
\newblock \emph{arXiv preprint arXiv:1912.02164}.

\bibitem[{Devlin et~al.(2019)Devlin, Chang, Lee, and
  Toutanova}]{devlin2019bert}
Jacob Devlin, Ming-Wei Chang, Kenton Lee, and Kristina Toutanova. 2019.
\newblock \href {https://doi.org/10.18653/v1/N19-1423} {{BERT}: Pre-training of
  deep bidirectional transformers for language understanding}.
\newblock In \emph{Proceedings of NAACL-HLT}.

\bibitem[{Dou et~al.(2021)Dou, Liu, Hayashi, Jiang, and Neubig}]{dou2021gsum}
Zi-Yi Dou, Pengfei Liu, Hiroaki Hayashi, Zhengbao Jiang, and Graham Neubig.
  2021.
\newblock \href {https://doi.org/10.18653/v1/2021.naacl-main.384} {{GS}um: A
  general framework for guided neural abstractive summarization}.
\newblock In \emph{Proceedings of NAACL-HLT 2021}.

\bibitem[{Dror et~al.(2018)Dror, Baumer, Shlomov, and
  Reichart}]{dror-etal-2018-hitchhikers}
Rotem Dror, Gili Baumer, Segev Shlomov, and Roi Reichart. 2018.
\newblock \href {https://doi.org/10.18653/v1/P18-1128} {The hitchhiker{'}s
  guide to testing statistical significance in natural language processing}.
\newblock In \emph{Proceedings of {ACL} 2018}.

\bibitem[{Ficler and Goldberg(2017)}]{ficler-goldberg-2017-controlling}
Jessica Ficler and Yoav Goldberg. 2017.
\newblock \href {https://doi.org/10.18653/v1/W17-4912} {Controlling linguistic
  style aspects in neural language generation}.
\newblock In \emph{Proceedings of {Style-Var} 2017}.

\bibitem[{Gu et~al.(2022)Gu, Feng, Ma, Zhang, Gong, and
  Qin}]{DBLP:journals/corr/abs-2210-02889}
Yuxuan Gu, Xiaocheng Feng, Sicheng Ma, Lingyuan Zhang, Heng Gong, and Bing Qin.
  2022.
\newblock \href {https://doi.org/10.48550/arXiv.2210.02889} {A distributional
  lens for multi-aspect controllable text generation}.
\newblock \emph{arXiv preprint arXiv:2210.02889}.

\bibitem[{He et~al.(2021)He, Zhou, Ma, Berg-Kirkpatrick, and
  Neubig}]{he2021towards}
Junxian He, Chunting Zhou, Xuezhe Ma, Taylor Berg-Kirkpatrick, and Graham
  Neubig. 2021.
\newblock \href {https://arxiv.org/abs/2110.04366} {Towards a unified view of
  parameter-efficient transfer learning}.
\newblock \emph{arXiv preprint arXiv:2110.04366}.

\bibitem[{He(2021)}]{DBLP:conf/emnlp/He21}
Xingwei He. 2021.
\newblock \href {https://doi.org/10.18653/v1/2021.emnlp-main.681} {Parallel
  refinements for lexically constrained text generation with {BART}}.
\newblock In \emph{Proceedings of {EMNLP} 2021}.

\bibitem[{Houlsby et~al.(2019)Houlsby, Giurgiu, Jastrzebski, Morrone,
  de~Laroussilhe, Gesmundo, Attariyan, and Gelly}]{houlsby2019parameter}
Neil Houlsby, Andrei Giurgiu, Stanislaw Jastrzebski, Bruna Morrone, Quentin
  de~Laroussilhe, Andrea Gesmundo, Mona Attariyan, and Sylvain Gelly. 2019.
\newblock \href {http://proceedings.mlr.press/v97/houlsby19a.html}
  {Parameter-efficient transfer learning for {NLP}}.
\newblock In \emph{Proceedings of ICML 2019}.

\bibitem[{Hu et~al.(2021)Hu, Shen, Wallis, Allen-Zhu, Li, Wang, Wang, and
  Chen}]{hu2021lora}
Edward~J Hu, Yelong Shen, Phillip Wallis, Zeyuan Allen-Zhu, Yuanzhi Li, Shean
  Wang, Lu~Wang, and Weizhu Chen. 2021.
\newblock \href {https://arxiv.org/abs/2106.09685} {Lora: Low-rank adaptation
  of large language models}.
\newblock \emph{arXiv preprint arXiv:2106.09685}.

\bibitem[{Huang et~al.(2021)Huang, Xu, Sun, and Liu}]{huang2021transfer}
Xuancheng Huang, Jingfang Xu, Maosong Sun, and Yang Liu. 2021.
\newblock \href {https://doi.org/10.18653/v1/2021.acl-long.446} {Transfer
  learning for sequence generation: from single-source to multi-source}.
\newblock In \emph{Proceedings of ACL-IJCNLP 2021}.

\bibitem[{Keskar et~al.(2019)Keskar, McCann, Varshney, Xiong, and
  Socher}]{keskar2019ctrl}
Nitish~Shirish Keskar, Bryan McCann, Lav~R Varshney, Caiming Xiong, and Richard
  Socher. 2019.
\newblock \href {https://arxiv.org/abs/1909.05858} {Ctrl: A conditional
  transformer language model for controllable generation}.
\newblock \emph{arXiv preprint arXiv:1909.05858}.

\bibitem[{Krause et~al.(2021)Krause, Gotmare, McCann, Keskar, Joty, Socher, and
  Rajani}]{krause-etal-2021-gedi-generative}
Ben Krause, Akhilesh~Deepak Gotmare, Bryan McCann, Nitish~Shirish Keskar,
  Shafiq Joty, Richard Socher, and Nazneen~Fatema Rajani. 2021.
\newblock \href {https://doi.org/10.18653/v1/2021.findings-emnlp.424}
  {{G}e{D}i: Generative discriminator guided sequence generation}.
\newblock In \emph{Findings of EMNLP 2021}.

\bibitem[{Kumar et~al.(2021)Kumar, Malmi, Severyn, and
  Tsvetkov}]{DBLP:conf/nips/KumarMST21}
Sachin Kumar, Eric Malmi, Aliaksei Severyn, and Yulia Tsvetkov. 2021.
\newblock \href
  {https://proceedings.neurips.cc/paper/2021/hash/79ec2a4246feb2126ecf43c4a4418002-Abstract.html}
  {Controlled text generation as continuous optimization with multiple
  constraints}.
\newblock In \emph{Proceedings of NeurIPS 2021}.

\bibitem[{Lample et~al.(2019)Lample, Subramanian, Smith, Denoyer, Ranzato, and
  Boureau}]{DBLP:conf/iclr/LampleSSDRB19}
Guillaume Lample, Sandeep Subramanian, Eric~Michael Smith, Ludovic Denoyer,
  Marc'Aurelio Ranzato, and Y{-}Lan Boureau. 2019.
\newblock \href {https://openreview.net/forum?id=H1g2NhC5KQ}
  {Multiple-attribute text rewriting}.
\newblock In \emph{Proceedings of {ICLR} 2019}.

\bibitem[{Lester et~al.(2021)Lester, Al-Rfou, and
  Constant}]{lester-etal-2021-power}
Brian Lester, Rami Al-Rfou, and Noah Constant. 2021.
\newblock \href {https://doi.org/10.18653/v1/2021.emnlp-main.243} {The power of
  scale for parameter-efficient prompt tuning}.
\newblock In \emph{Proceedings of EMNLP 2021}.

\bibitem[{Lewis et~al.(2020)Lewis, Liu, Goyal, Ghazvininejad, Mohamed, Levy,
  Stoyanov, and Zettlemoyer}]{lewis2020bart}
Mike Lewis, Yinhan Liu, Naman Goyal, Marjan Ghazvininejad, Abdelrahman Mohamed,
  Omer Levy, Veselin Stoyanov, and Luke Zettlemoyer. 2020.
\newblock \href {https://doi.org/10.18653/v1/2020.acl-main.703} {{BART}:
  Denoising sequence-to-sequence pre-training for natural language generation,
  translation, and comprehension}.
\newblock In \emph{Proceedings of ACL 2020}.

\bibitem[{Li et~al.(2016)Li, Galley, Brockett, Gao, and
  Dolan}]{li-etal-2016-diversity}
Jiwei Li, Michel Galley, Chris Brockett, Jianfeng Gao, and Bill Dolan. 2016.
\newblock \href {https://doi.org/10.18653/v1/N16-1014} {A diversity-promoting
  objective function for neural conversation models}.
\newblock In \emph{Proceedings of {NAACL-HLT} 2016}.

\bibitem[{Li and Liang(2021)}]{li-liang-2021-prefix}
Xiang~Lisa Li and Percy Liang. 2021.
\newblock \href {https://doi.org/10.18653/v1/2021.acl-long.353} {Prefix-tuning:
  Optimizing continuous prompts for generation}.
\newblock In \emph{Proceedings of ACL-IJCNLP 2021}.

\bibitem[{Lin and Riedl(2021)}]{lin-riedl-2021-plug}
Zhiyu Lin and Mark Riedl. 2021.
\newblock \href {https://doi.org/10.18653/v1/2021.nuse-1.7} {Plug-and-blend: A
  framework for controllable story generation with blended control codes}.
\newblock In \emph{Proceedings of NUSE 2021}.

\bibitem[{Liu et~al.(2020)Liu, Gu, Goyal, Li, Edunov, Ghazvininejad, Lewis, and
  Zettlemoyer}]{liu2020multilingual}
Yinhan Liu, Jiatao Gu, Naman Goyal, Xian Li, Sergey Edunov, Marjan
  Ghazvininejad, Mike Lewis, and Luke Zettlemoyer. 2020.
\newblock \href {https://doi.org/10.1162/tacl_a_00343} {Multilingual denoising
  pre-training for neural machine translation}.
\newblock \emph{TACL}.

\bibitem[{Liu et~al.(2019)Liu, Ott, Goyal, Du, Joshi, Chen, Levy, Lewis,
  Zettlemoyer, and Stoyanov}]{liu2019roberta}
Yinhan Liu, Myle Ott, Naman Goyal, Jingfei Du, Mandar Joshi, Danqi Chen, Omer
  Levy, Mike Lewis, Luke Zettlemoyer, and Veselin Stoyanov. 2019.
\newblock \href {https://arxiv.org/abs/1907.11692} {Roberta: A robustly
  optimized bert pretraining approach}.
\newblock \emph{arXiv preprint arXiv:1907.11692}.

\bibitem[{Lyu et~al.(2021)Lyu, Liang, Pham, Hovy, P{\'o}czos, Salakhutdinov,
  and Morency}]{lyu-etal-2021-styleptb}
Yiwei Lyu, Paul~Pu Liang, Hai Pham, Eduard Hovy, Barnab{\'a}s P{\'o}czos,
  Ruslan Salakhutdinov, and Louis-Philippe Morency. 2021.
\newblock \href {https://doi.org/10.18653/v1/2021.naacl-main.171}
  {{S}tyle{PTB}: A compositional benchmark for fine-grained controllable text
  style transfer}.
\newblock In \emph{Proceedings of {NAACL-HLT} 2021}.

\bibitem[{Nallapati et~al.(2016)Nallapati, Zhou, dos Santos, Gu̇l{\c{c}}ehre,
  and Xiang}]{nallapati-etal-2016-abstractive}
Ramesh Nallapati, Bowen Zhou, Cicero dos Santos, {\c{C}}a{\u{g}}lar
  Gu̇l{\c{c}}ehre, and Bing Xiang. 2016.
\newblock \href {https://doi.org/10.18653/v1/K16-1028} {Abstractive text
  summarization using sequence-to-sequence {RNN}s and beyond}.
\newblock In \emph{Proceedings of {SIGNLL} 2016}.

\bibitem[{Nguyen and Verspoor(2018)}]{nguyen-verspoor-2018-improved}
Dat~Quoc Nguyen and Karin Verspoor. 2018.
\newblock \href {https://doi.org/10.18653/v1/K18-2008} {An improved neural
  network model for joint {POS} tagging and dependency parsing}.
\newblock In \emph{Proceedings of {C}o{NLL} 2018}.

\bibitem[{Papineni et~al.(2002)Papineni, Roukos, Ward, and
  Zhu}]{papineni-etal-2002-bleu}
Kishore Papineni, Salim Roukos, Todd Ward, and Wei-Jing Zhu. 2002.
\newblock \href {https://doi.org/10.3115/1073083.1073135} {{B}leu: a method for
  automatic evaluation of machine translation}.
\newblock In \emph{Proceedings of {ACL} 2002}.

\bibitem[{Post(2018)}]{post2018call}
Matt Post. 2018.
\newblock \href {https://doi.org/10.18653/v1/W18-6319} {A call for clarity in
  reporting {BLEU} scores}.
\newblock In \emph{Proceedings of WMT 2018}.

\bibitem[{Post and Vilar(2018)}]{post-vilar-2018-fast}
Matt Post and David Vilar. 2018.
\newblock \href {https://doi.org/10.18653/v1/N18-1119} {Fast lexically
  constrained decoding with dynamic beam allocation for neural machine
  translation}.
\newblock In \emph{Proceedings of NAACL 2018}.

\bibitem[{Qian et~al.(2022)Qian, Dong, Shen, Wei, and
  Chen}]{qian-etal-2022-controllable}
Jing Qian, Li~Dong, Yelong Shen, Furu Wei, and Weizhu Chen. 2022.
\newblock \href {https://doi.org/10.18653/v1/2022.findings-acl.229}
  {Controllable natural language generation with contrastive prefixes}.
\newblock In \emph{Findings of ACL 2022}.

\bibitem[{Qin et~al.(2022)Qin, Welleck, Khashabi, and
  Choi}]{DBLP:journals/corr/abs-2202-11705}
Lianhui Qin, Sean Welleck, Daniel Khashabi, and Yejin Choi. 2022.
\newblock \href {https://arxiv.org/abs/2202.11705} {{COLD} decoding:
  Energy-based constrained text generation with langevin dynamics}.
\newblock \emph{arXiv preprint arXiv:2202.11705}.

\bibitem[{Radford et~al.(2019)Radford, Wu, Child, Luan, Amodei, and
  Sutskever}]{radford2019language}
Alec Radford, Jeffrey Wu, Rewon Child, David Luan, Dario Amodei, and Ilya
  Sutskever. 2019.
\newblock \href {http://www.persagen.com/files/misc/radford2019language.pdf}
  {Language models are unsupervised multitask learners}.
\newblock \emph{OpenAI blog}.

\bibitem[{Tan et~al.(2020)Tan, Zhang, Huang, Chen, Wang, Sun, Luan, and
  Liu}]{tan-etal-2020-thumt}
Zhixing Tan, Jiacheng Zhang, Xuancheng Huang, Gang Chen, Shuo Wang, Maosong
  Sun, Huanbo Luan, and Yang Liu. 2020.
\newblock \href {https://aclanthology.org/2020.amta-research.11} {{THUMT}: An
  open-source toolkit for neural machine translation}.
\newblock In \emph{Proceedings of {AMTA} 2020}.

\bibitem[{Vaswani et~al.(2017)Vaswani, Shazeer, Parmar, Uszkoreit, Jones,
  Gomez, Kaiser, and Polosukhin}]{vaswani2017attention}
Ashish Vaswani, Noam Shazeer, Niki Parmar, Jakob Uszkoreit, Llion Jones,
  Aidan~N. Gomez, Lukasz Kaiser, and Illia Polosukhin. 2017.
\newblock \href
  {https://proceedings.neurips.cc/paper/2017/hash/3f5ee243547dee91fbd053c1c4a845aa-Abstract.html}
  {Attention is all you need}.
\newblock In \emph{Proceedings of NeurIPS 2017}.

\bibitem[{Wang et~al.(2022)Wang, Tan, and Liu}]{DBLP:conf/acl/WangTL22}
Shuo Wang, Zhixing Tan, and Yang Liu. 2022.
\newblock \href {https://doi.org/10.18653/v1/2022.acl-long.487} {Integrating
  vectorized lexical constraints for neural machine translation}.
\newblock In \emph{Proceedings of {ACL} 2022}.

\bibitem[{Yang and Klein(2021)}]{DBLP:conf/naacl/YangK21}
Kevin Yang and Dan Klein. 2021.
\newblock \href {https://doi.org/10.18653/v1/2021.naacl-main.276} {{FUDGE:}
  controlled text generation with future discriminators}.
\newblock In \emph{Proceedings of {NAACL-HLT} 2021}.

\bibitem[{Yang et~al.(2022)Yang, Liu, Lei, Yang, Xue, Chen, and
  Xie}]{DBLP:journals/corr/abs-2204-13362}
Kexin Yang, Dayiheng Liu, Wenqiang Lei, Baosong Yang, Mingfeng Xue, Boxing
  Chen, and Jun Xie. 2022.
\newblock \href {https://doi.org/10.48550/arXiv.2204.13362} {Tailor: {A}
  prompt-based approach to attribute-based controlled text generation}.
\newblock \emph{arXiv preprint arXiv:2204.13362}.

\bibitem[{Zoph and Knight(2016)}]{zoph2016multi}
Barret Zoph and Kevin Knight. 2016.
\newblock \href {https://doi.org/10.18653/v1/N16-1004} {Multi-source neural
  translation}.
\newblock In \emph{Proceedings of NAACL-HLT 2016}.

\end{thebibliography}

\bibliographystyle{acl_natbib}

\clearpage

\appendix

\section{Theoretical Analysis} \label{app:theory}

In this section, we theoretically analyze mutual interference (MI) and derive a lower bound of MI for prefix-tuning~\cite{li-liang-2021-prefix}. As Feed Forward and Layernorm sublayers are position-wise operations~\cite{vaswani2017attention} which would not introduce the interference of plugins, we focus on analyzing the multi-head attention (MHA) sublayers.

According to the previous study~\cite{he2021towards}, the output of a single head of attention with prefixes of the $i$-th plugin, which is represented by $\mathbf{h}_i$, could be described as
\begin{alignat}{1}
\mathbf{h}_i &= \lambda(\mathbf{x}_i) \overline{\mathbf{h}}_i  + \big(1-\lambda(\mathbf{x}_i)\big)\Delta \mathbf{h}_i \nonumber \\
 &= s_i\overline{\mathbf{h}}_i + t_i\Delta \mathbf{h}_i, \label{eqn:app8}
\end{alignat}
where $\overline{\mathbf{h}}_i$ denotes the original output of the pretrained generative model with $\mathbf{x}_i$ as input. $\lambda(\mathbf{x}_i)$ is a scalar related to the attention weights, where $\lambda(\mathbf{x}_i)=s_i=1-t_i \in (0,1)$. In addition, $\Delta \mathbf{h}_i$ is an offset determined by the $i$-th plugin, and its magnitude is positively correlated with the magnitude of $\bm\phi_i$, where $\bm\phi_i$ is the set of parameters of the $i$-th plugin.

Following the pattern above, when the $i$-th and $j$-th plugins are inserted at the same time, the output of the head (i.e., $\mathbf{h}_{i,j}$) turns to be
\begin{alignat}{1}
\mathbf{h}_{i,j} = \gamma\overline{\mathbf{h}}_{i,j} + \alpha \Delta{\mathbf{h}_i} + \beta \Delta{\mathbf{h}_j},
\end{alignat}
where $\overline{\mathbf{h}}_{i,j}$ is the output of pretrained generative model, and $\alpha,\beta,\gamma \in (0,1), \alpha < t_i, \beta < t_j, \gamma < s_i, \gamma < s_j$. Similarly, $\Delta {\mathbf{h}_i}$ and $\Delta{\mathbf{h}_j}$ are determined by the $i$-th and $j$-th plugins.

According to the definition in Eq.~(\ref{eqn:definition}), let $\tilde{\mathbf{h}}_{i,j}$ and $\hat{\mathbf{h}}_{i,j}$ be the outputs like $\mathbf{h}_{i,j}$ after training on multi- and single-aspect labeled data, respectively. The mutual interference of two plugins in a single head (i.e., $\mathrm{MI}_s$) can be measured by the norm of the gap between outputs under supervised and zero-shot inference:
\begin{alignat}{1}
\mathrm{MI}_s &=  \big\Vert\tilde{\mathbf{h}}_{i,j} - \hat{\mathbf{h}}_{i,j}\big\Vert \nonumber \\
&= \big\Vert \tilde{\mathbf{h}}_{i,j} - (\gamma\overline{\mathbf{h}}_{i,j} + \alpha \Delta \hat{\mathbf{h}}_i + \beta \Delta \hat{\mathbf{h}}_j) \big\Vert \nonumber \\
&\ge \big\Vert \tilde{\mathbf{h}}_{i,j} - \gamma \overline{\mathbf{h}}_{i,j}\big\Vert - \big\Vert\alpha \Delta \hat{\mathbf{h}}_i + \beta \Delta \hat{\mathbf{h}}_j \big\Vert,
\end{alignat}
where $\Delta \hat{\mathbf{h}}_i$ and $\Delta \hat{\mathbf{h}}_j$ correspond to offsets that plugins are trained on single-aspect labeled data.

Considering that the intervention caused by two plugins simultaneously should larger than the sum of two interventions caused by two plugins respectively because of the interaction between two plugins, we assume that there is 
\begin{alignat}{1}
\big\Vert \tilde{\mathbf{h}}_{i,j} - \overline{\mathbf{h}}_{i,j}\big\Vert > \big\Vert \hat{\mathbf{h}}_{i} - \overline{\mathbf{h}}_i\big\Vert + \big\Vert \hat{\mathbf{h}}_{j} - \overline{\mathbf{h}}_j\big\Vert .
\end{alignat}
Based on this, we can derive 
\begin{alignat}{1}
 \mathrm{MI}_s > & \big\Vert \hat{\mathbf{h}}_i - \gamma\overline{\mathbf{h}}_i \big\Vert + \big\Vert \hat{\mathbf{h}}_j -\gamma\overline{\mathbf{h}}_j \big\Vert \nonumber \\
 & - \big\Vert\alpha \Delta \hat{\mathbf{h}}_i + \beta \Delta \hat{\mathbf{h}}_j \big\Vert.
\end{alignat}
Given that $ s_i > \gamma, s_j > \gamma $, and $\hat{\mathbf{h}}_i=s_i\overline{\mathbf{h}}_i + t_i\Delta \hat{\mathbf{h}}_i$ (Eq.~(\ref{eqn:app8})), $\mathrm{MI}_s$ satisfies
\begin{alignat}{1}
\mathrm{MI}_s >&\;  \big\Vert \hat{\mathbf{h}}_i - s_i\overline{\mathbf{h}}_i \big\Vert + \big\Vert \hat{\mathbf{h}}_j -s_j\overline{\mathbf{h}}_j \big\Vert \nonumber \\ 
 &\; - \big\Vert\alpha \Delta \hat{\mathbf{h}}_i + \beta \Delta \hat{\mathbf{h}}_j \big\Vert \nonumber \\
= &\; \Vert t_i \Delta \hat{\mathbf{h}}_i\Vert + \Vert t_j \Delta \hat{\mathbf{h}}_j\Vert - \Vert\alpha \Delta \hat{\mathbf{h}}_i + \beta \Delta \hat{\mathbf{h}}_j \Vert \nonumber \\
\ge &\; (t_i-\alpha)  \Vert\Delta \hat{\mathbf{h}}_i\Vert + (t_j-\beta) \Vert\Delta \hat{\mathbf{h}}_j\Vert, \label{eqn:mi_singlehead}
\end{alignat}
where $ 1 > t_i-\alpha > 0$ and $1 > t_j-\beta>0$. Therefore, the mutual interference of two plugins in a single head has a positive lower bound, and it is positively correlated with the magnitude of $\hat{\bm{\phi}}_i$ and $\hat{\bm{\phi}}_j$.

To step further, we derive the lower bound of MI in the multi-head scenario. Assume that $K$ denotes the number of heads, $\mathbf{W}_o$ denotes the fixed output projection matrix in the MHA, $\mathbf{W}_o = \mathbf{Q}_o\mathbf{R}_o$ is the QR-decomposition format of $\mathbf{W}_o$, $\hat{\lambda_o}$ is the average of absolute eigenvalues. Specifically, $\hat{\mathbf{h}}^k_{i,j}$ and $\tilde{\mathbf{h}}^k_{i,j}$ denotes $\hat{\mathbf{h}}_{i,j}$ and $\tilde{\mathbf{h}}_{i,j}$ in the $k$-th head, respectively.

Then, the lower bound of MI in MHA (i.e., $\mathrm{MI}_m$) can be derived as (viewing $\mathbf{R}_o$ as a diagonal matrix for simplicity)
\begin{alignat}{1}
 \mathrm{MI}_m &= \big\Vert\mathrm{concat}(\tilde{\mathbf{h}}^k_{i,j} -\hat{\mathbf{h}}^k_{i,j})_{k=1}^K \mathbf{W}_o \big\Vert \nonumber \\
 &= \big\Vert\mathrm{concat}(\tilde{\mathbf{h}}^k_{i,j}-\hat{\mathbf{h}}^k_{i,j})_{k=1}^K \mathbf{Q}_o \mathbf{R}_o \big\Vert \nonumber \\
 &\approx  \hat{\lambda_o} \big\Vert\mathrm{concat}(\tilde{\mathbf{h}}^k_{i,j}-\hat{\mathbf{h}}^k_{i,j})_{k=1}^K \mathbf{Q}_o\big\Vert \nonumber \\
 &= \hat{\lambda_o} \sqrt{\sum _{k=1} ^K \big\Vert \tilde{\mathbf{h}}^k_{i,j}-\hat{\mathbf{h}}^k_{i,j} \big\Vert^2} \nonumber \\
 &\ge \frac{\hat{\lambda_o}}{\sqrt{n}} \sum_{k=1}^K \big\Vert \tilde{\mathbf{h}}^k_{i,j}-\hat{\mathbf{h}}^k_{i,j} \big\Vert \nonumber \\
 &> \frac{\hat{\lambda_o}}{\sqrt{n}}  \sum_{k=1}^K \Big((t_i^k-\alpha^k)  \Vert\Delta \hat{\mathbf{h}}_{i}^k\Vert \nonumber \\
 &+ (t_j^k-\beta^k) \Vert\Delta \hat{\mathbf{h}}_{j}^k\Vert\Big), \label{eqn:mi_multihead}
\end{alignat}
where $ 1 > t_i^k-\alpha^k > 0$ and $1 > t_j^k-\beta^k>0$, and $\Delta \hat{\mathbf{h}}_{i}^k$ and $\Delta \hat{\mathbf{h}}_{j}^k$ are also positively correlated with the magnitude of $\hat{\bm{\phi}}_i$ and $\hat{\bm{\phi}}_j$, respectively. 

Therefore, the mutual interference of multiple plugins has a theoretical positive lower bound, which implies concatenating prefixes that are separately trained has an irreparable gap against supervised-trained prefixes. As a result, MI might accumulate along with the depth of the model, like in Figure~\ref{fig:interference}.
Intuitively, introducing gates, which contain trainable coefficients between 0 to 1, to $\hat{\bm{\phi}}_i$ is helpful for decreasing the offsets in Eq.~(\ref{eqn:mi_multihead}) and thus mutual interference.

\section{Reproducibility} \label{app:model_setup}

\subsection{Data Preparation}

For text generation, we adopted the widely-used Yelp dataset\footnote{\url{https://github.com/shrimai/Style-Transfer-Through-Back-Translation}}~\cite{DBLP:conf/iclr/LampleSSDRB19}, which contains restaurant reviews with sentiment (positive and negative) and topic (American, Mexican, and Asian) labels. Specifically, following previous work~\cite{DBLP:journals/corr/abs-2204-13362}, we randomly sampled 30K/3K sentences for each attribute for training/validation while ensuring the balance of different attributes in the final dataset (Table \ref{tab:data}). For evaluation, we sampled 25 sentences for each given textual prefix and combination of aspects. In addition, we eliminated the sentences rated 3 in sentiment. To evaluate the extensibility of methods, we added two additional aspects of constraints: keywords~\cite{DBLP:conf/emnlp/He21} and tense (past and present)~\cite{ficler-goldberg-2017-controlling}, where their labels are automatically extracted from the reviews. More precisely, we randomly extracted 1 to 3 words as keywords for each sentence~\cite{post-vilar-2018-fast}, and the tenses of sentences are labeled by an open-source toolkit\footnote{\url{https://github.com/ajitrajasekharan/simple_tense_detector}} that is based on a POS tagger~\cite{nguyen-verspoor-2018-improved}.

For machine translation, we adopted the WMT14 German $\rightarrow$ English benchmark\footnote{\url{https://statmt.org/wmt14/translation-task.html}}. Specifically, the training, validation, and test sets contain 4,500K, 3K, and 3K sentences, respectively. We adopted three aspects of constraints in machine translation, and they are all automatically obtained from target sentences. We use keywords~\cite{post-vilar-2018-fast} and tense~\cite{ficler-goldberg-2017-controlling} like the text generation task to control translations. For the third constraint, we adopt French synonymous sentences as external knowledge, which is beneficial to disambiguation. Note that it does not directly control any attribute of translations but will improve the translation quality~\cite{zoph2016multi}. The French synonymous sentences are given by a Transformer-based English$\rightarrow$French translation model~\cite{vaswani2017attention}. 

 \begin{table}[!t]
    \begin{center}
    \resizebox{\linewidth}{!}{
        \begin{tabular}{l|ccc}
        \toprule
        {\bf Task} & {\bf Training} & {\bf Validation} & {\bf Test} \\ \midrule \midrule
        {Text Generation} & \ \ \ \ \   30K & 3K & 375$^*$ \\
        {Machine Translation} & 4,500K & 3K & 3K \\
        \bottomrule
        \end{tabular}
        }
    \caption{The number of examples (sentences) in Training/Validation/Test for each attribute in Text Generation and aspect in Machine Translation. *: For Test in Text Generation, to keep in line with previous works~\cite{DBLP:journals/corr/abs-2204-13362}, we use 15 attribute-unrelated prefixes (as listed in \S\ref{app:eval_metric}) and ask model to continue writing under each attribute (25 sentences for each). }
    \label{tab:data}
    \end{center}
\end{table}

\subsection{Evaluation Metrics}\label{app:eval_metric} 

For text generation, following previous work~\cite{lyu-etal-2021-styleptb, dathathri2019plug, DBLP:journals/corr/abs-2210-02889}, we adopted automatic and human evaluation for constraint accuracy and text quality. For the evaluation of sentiment and topic, we finetuned two RoBERTa-based~\cite{liu2019roberta} classifiers on the Yelp dataset. Specifically, we randomly over-sampled 1,500K/15K/15K sentences for training/validation/test set of topic and 1,380K/1K/1K sentences for training/validation/test set of sentiment. The F1 scores for sentiment and topic are 98.71 and 89.62, respectively. The same toolkit as training evaluated the accuracy of tense, and we used word-level Copy Success Rate (CSR)~\cite{chen2021lexical} to evaluate the lexical constraint. In addition, we used the perplexity (PPL) given by GPT-2$_\mathrm{medium}$~\cite{radford2019language} and averaged distinctness~\cite{li-etal-2016-diversity} to evaluate the fluency and diversity of the generated text, respectively. Similar to previous work~\cite{dathathri2019plug,DBLP:conf/naacl/YangK21}, we used 15 textual prefixes\footnote{\url{https://github.com/uber-research/PPLM}} and asked models to start writing from them for each combination of constraints during inference. For human evaluation, each sentence received a score of 1 to 5 on sentiment and topic relevance as well as fluency given by three evaluators. The final scores are averaged over three ratings. 

Specifically, the 15 textual prefixes are: ``Once upon a time'',``The book'',``The chicken'',``The city'',``The country'',``The lake'',``The movie'',``The painting'',``The weather'',``The food'',``While this is happening'',``The pizza'',``The potato'',``The president of the country'',``The year is 1910.''.

For machine translation, we adopted \textsc{SacreBLEU}\footnote{The signature is ``BLEU+case.mixed+numrefs.1+smooth .exp+tok.13a+version.2.0.0''.} \cite{post2018call} to evaluate the translation quality. Similar to text generation, we used CSR~\cite{chen2021lexical} and tense accuracy to evaluate lexical and tense constraints, respectively.

\begin{table}[!t]
    \begin{center}
        \scalebox{0.8}{
        \begin{tabular}{lrr}
        \toprule
        \bf Hyper-parameter & \bf TG & \bf MT \\
        \midrule \midrule
        \multicolumn{3}{c}{\em Pretrained Model} \\
        \midrule
                        Encoder layers & $12$ & $12$ \\
                        Decoder layers & $12$ & $12$ \\
                        Attention heads & $16$ & $16$ \\
                        Attention head size & $64$ & $64$ \\
                        Hidden size & $1,024$ & $1,024$ \\
                        FFN hidden size & $4,096$ & $4,096$ \\
                        Max sentence length & $1,024$ & $1,024$ \\
        \midrule
        \multicolumn{3}{c}{\em Training} \\
        \midrule 
                        Optimizer & Adam & Adam \\
                        Adam beta & (0.9, 0.999) & (0.9, 0.999) \\
                        Training steps & $50,000$ & $150,000$ \\
                        Warmup steps & $10,000$ & $10,000$ \\
                        Batch size & $1,024$ & $512$ \\
                        Learning rate & $1\times 10^{-4}$ & $4\times 10^{-4}$ \\
                        Initial learning rate & $5\times 10^{-8}$ & $5\times 10^{-8}$ \\
                        Residual dropout & $0.1$ & $0.1$ \\
                        Attention dropout & $0.0$ & $0.0$ \\
                        Activation dropout & $0.0$ & $0.0$ \\
        \midrule
        \multicolumn{3}{c}{\em Inference} \\
        \midrule
                        Length penalty & $0.6$ & $1.2$ \\
                        Top $K$ & $10$ & / \\
                        Beam size & / & $5$ \\
        \bottomrule
        \end{tabular}
        }
        \caption{The commonly-used hyper-parameters in text generation (TG) and machine translation (MT).}
        \label{tab:common_hyper_para}
    \end{center}
\end{table}

\begin{table}[!t]
    \begin{center}
    \resizebox{\linewidth}{!}{
        \begin{tabular}{l|r}
        \toprule
        {\bf Method}& {\bf{\# Trainable Parameters}} \\ \midrule \midrule
        {\textsc{Gedi}}&  $345M$ \\
        {\textsc{Dist. Lens}}& $110M + 768^2 + 768\times2\times20\times1024\times24 = 866M $ \\
        \textsc{Parallel Adapter}& $2 \times 19 \times 1024 \times (36 + 24) \approx 2.33M$ \\ 
        {\textsc{LoRA}} & $4 \times 17 \times 1024 \times 36 \approx 2.51M $ \\
        {\textsc{Prompt-tuning}} & $100 \times 1024 \approx 0.10M$ \\ 
        {\textsc{Prefix-tuning}} & $2 \times 33 \times 1024 \times 36 \approx 2.43M $ \\
        {\textsc{Tailor}}& $100 \times 1024 \approx 0.10M$  \\
        {\textsc{Prompt Gating}} & $1024 \times 100 \times 25 \approx 2.56M $ \\
        \bottomrule
        \end{tabular}
        }
    \caption{The number of trainable parameters of a single aspect for each method.}
    \label{tab:num_parameters}
    \end{center}
\end{table}

\subsection{Model and Hyper-parameters}

As our approach has both encoder and decoder, we adopted BART-large\footnote{\url{https://huggingface.co/facebook/bart-large}}~\cite{lewis2020bart} for text generation and mBART-large-cc25\footnote{\url{https://huggingface.co/facebook/mbart-large-cc25}}~\cite{liu2020multilingual} for machine translation.

\begin{figure}[!t]
  \centering
  \includegraphics[width=.49\linewidth]{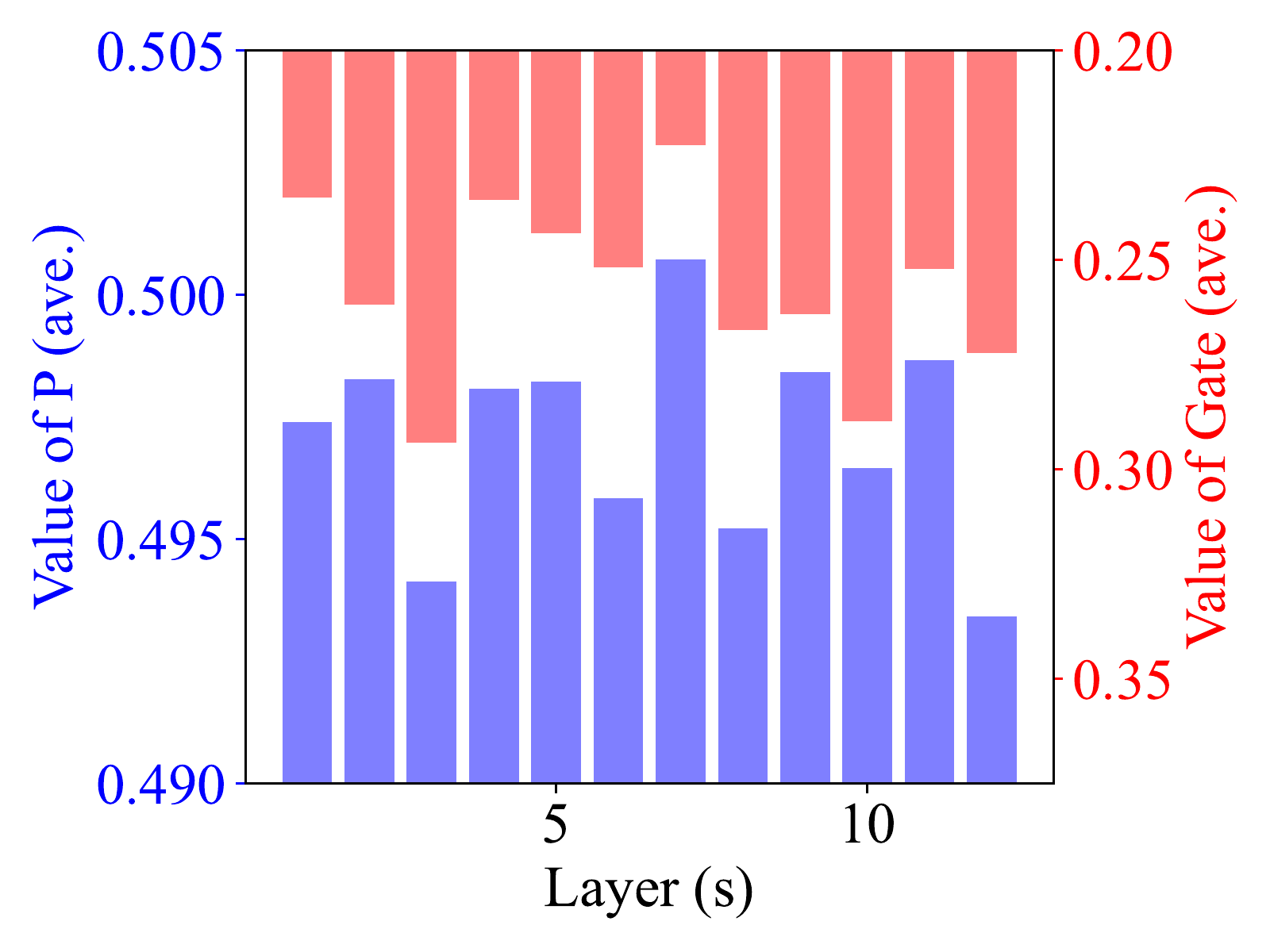}
  \includegraphics[width=.49\linewidth]{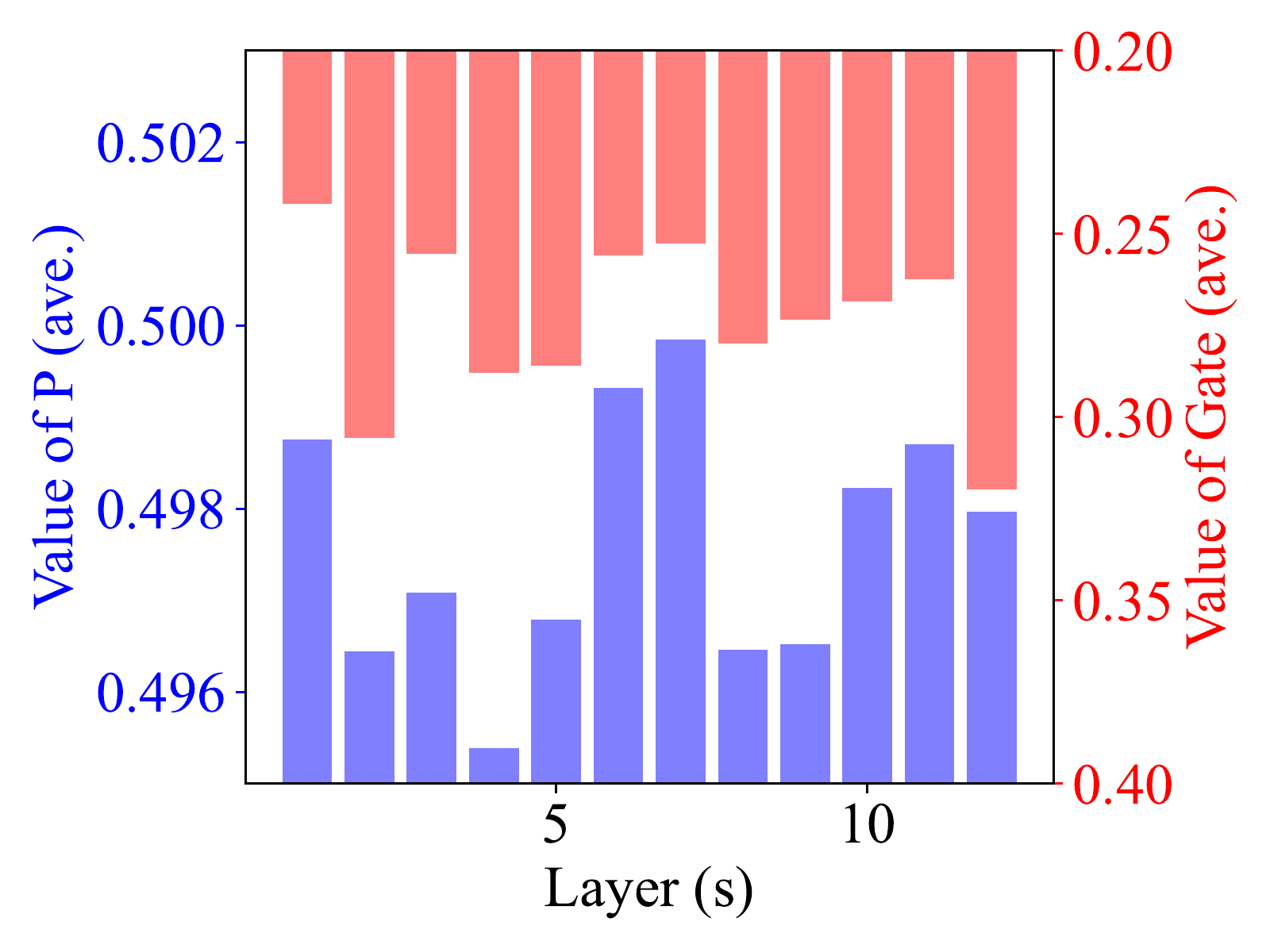}
  \includegraphics[width=.49\linewidth]{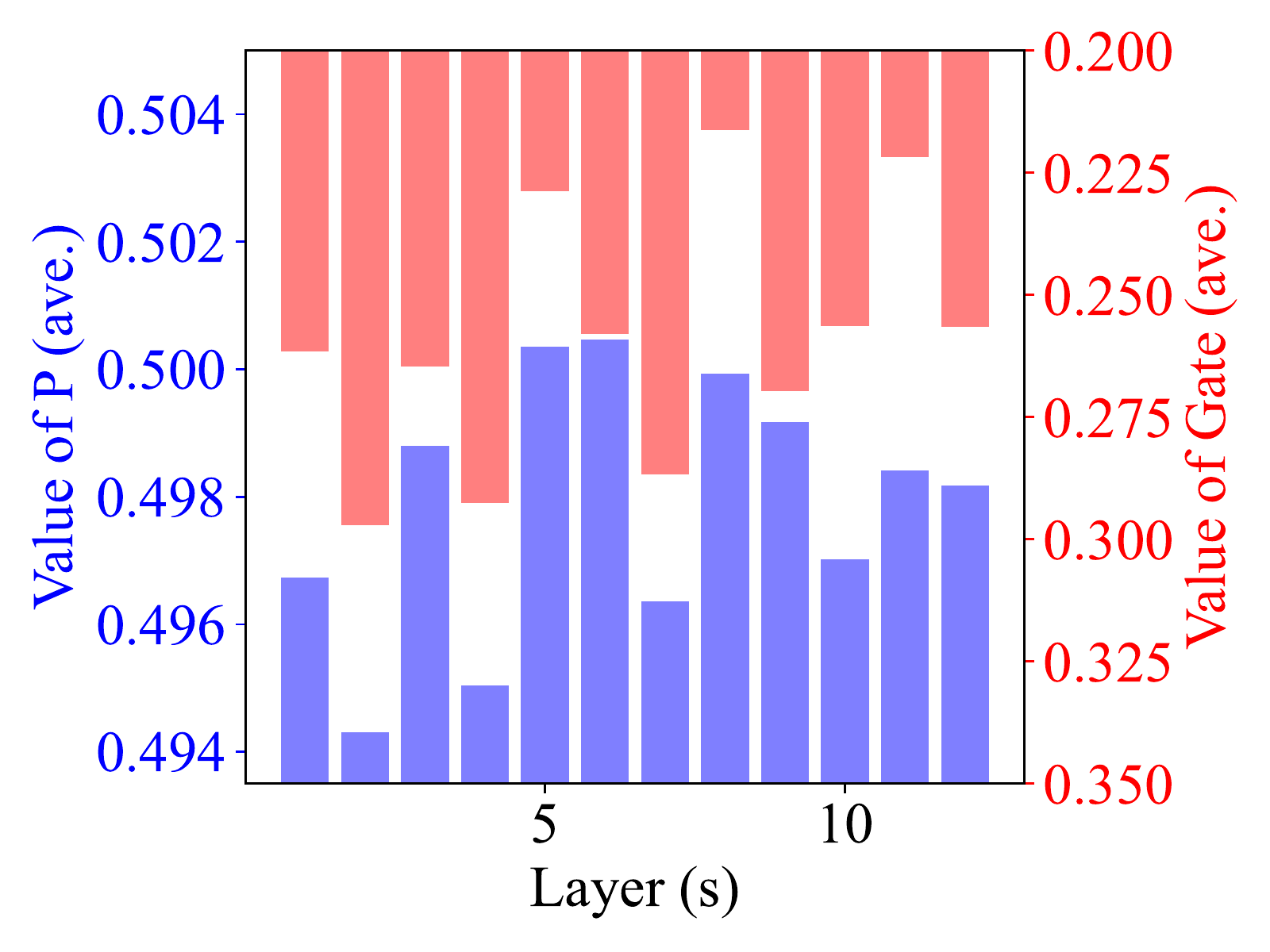}
  \includegraphics[width=.49\linewidth]{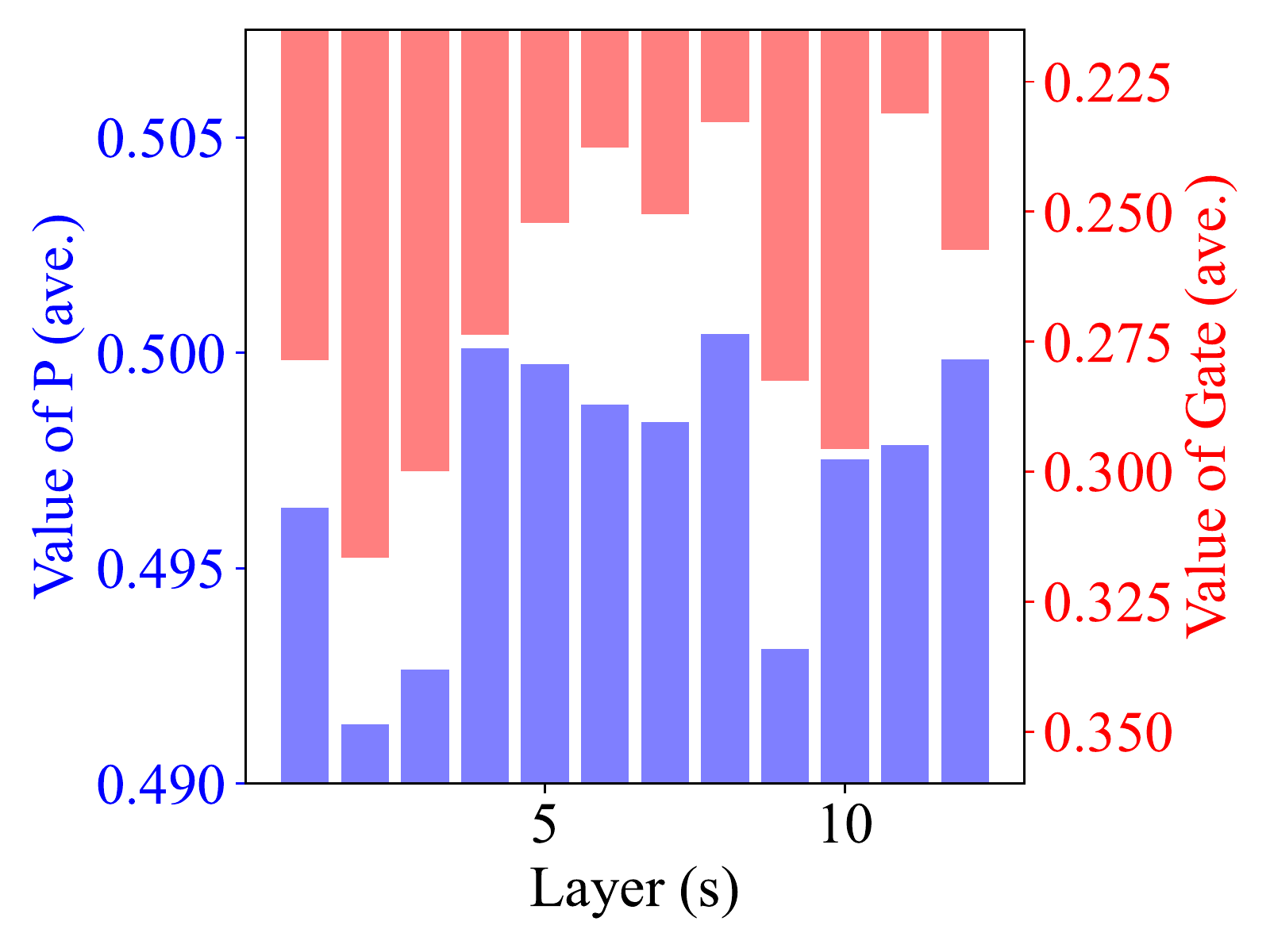}
  \caption{
  The visualization on text generation. The average value of gates $\sigma (\textbf{G}_i^{(j)})$ (red bars) and the average of $L_1$ norm of $\textbf{P}_i^{(j)}$ (blue bars) on each layer, according to Eq.~(\ref{eqn:gate}). The values are extracted for the sentiment aspect, including negative (top left) and positive (top right), and topic aspect, including Asian (bottom left) and American (bottom right).
  }
  \label{fig:stability}
\end{figure}

\begin{figure}[!t]
  \centering
  \includegraphics[width=.49\linewidth]{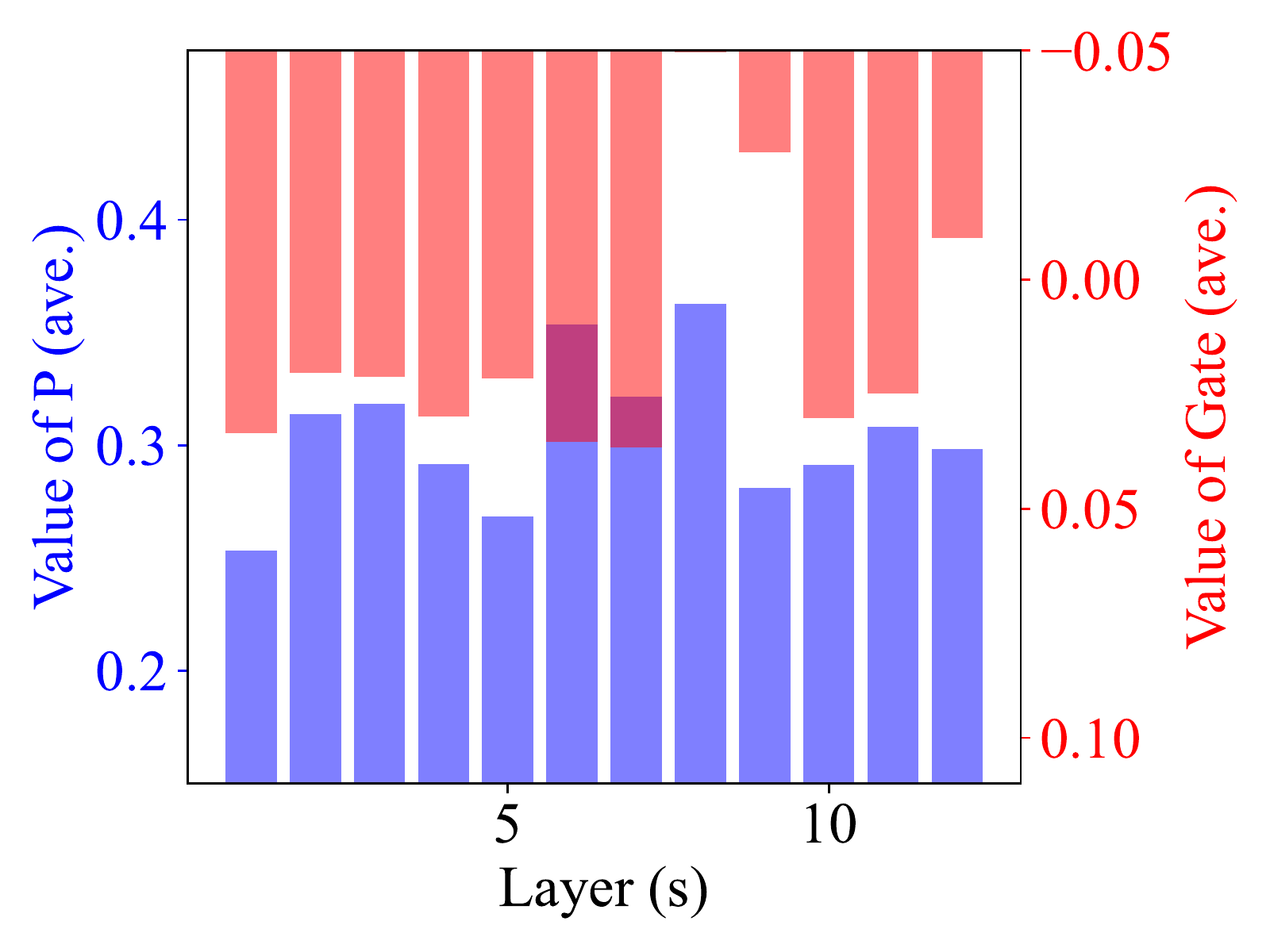}
  \includegraphics[width=.49\linewidth]{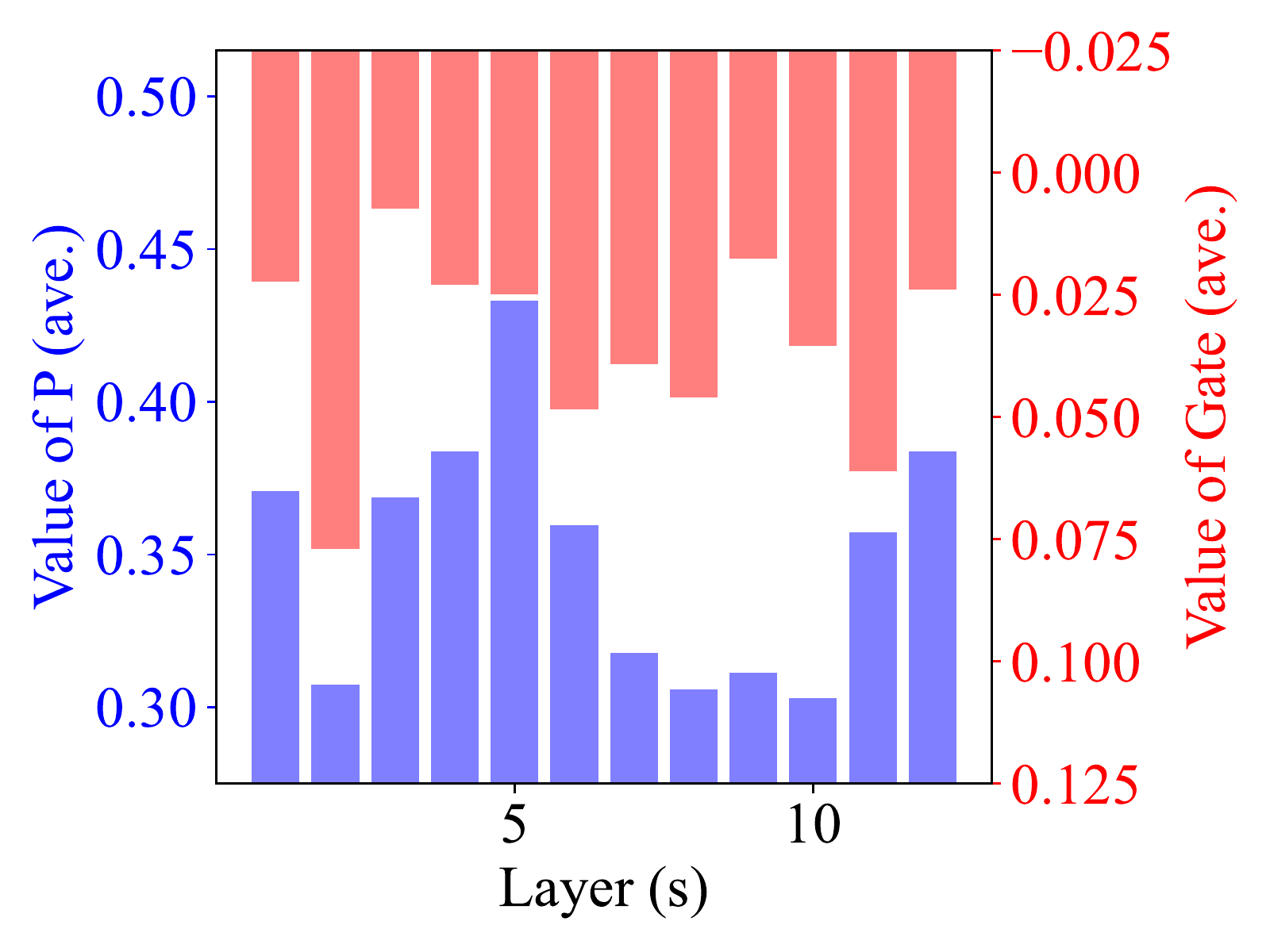}
  \includegraphics[width=.49\linewidth]{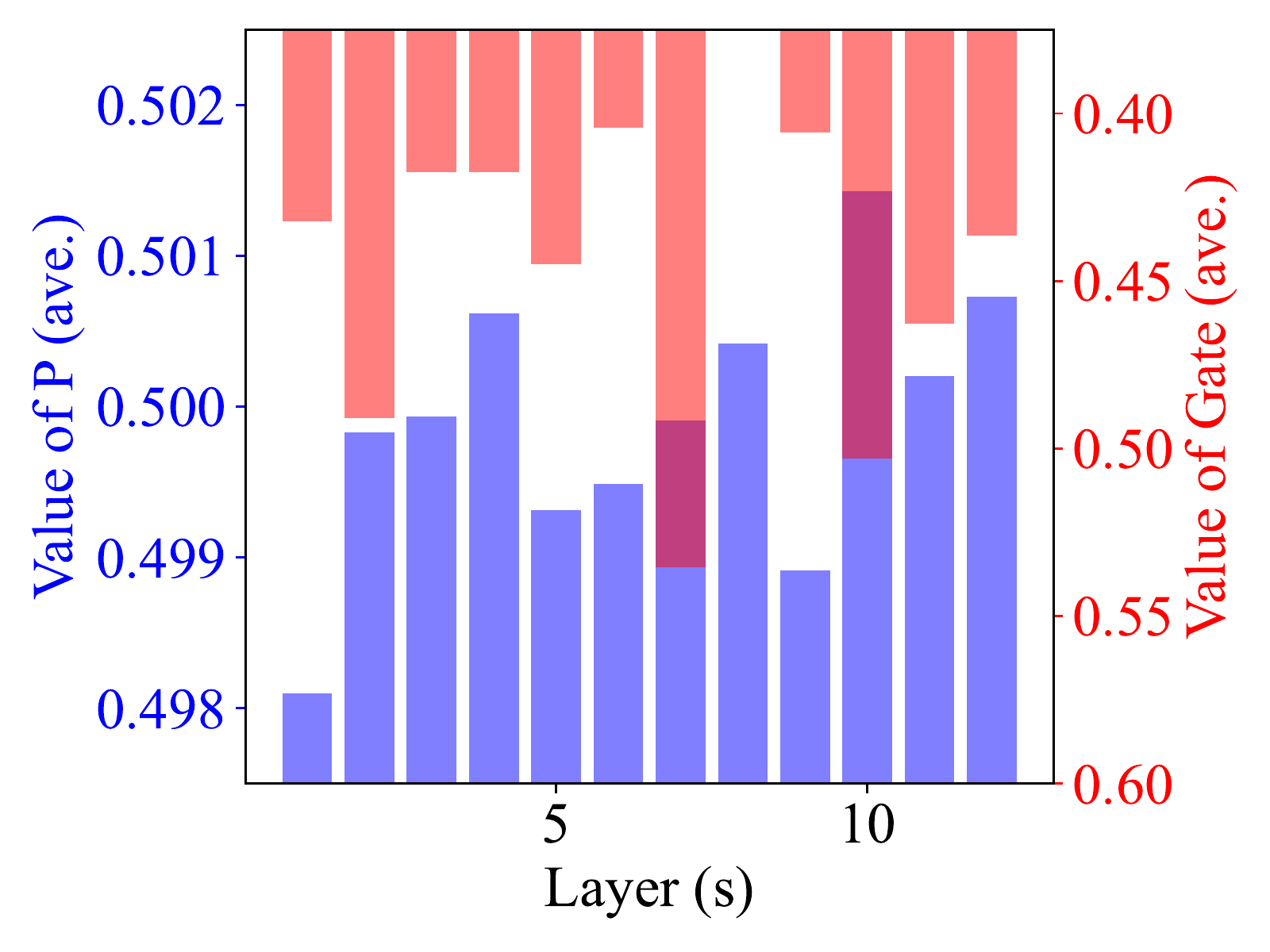}

  \caption{
  The visualization on machine translation. The average value of gates $\sigma (\textbf{G}_i^{(j)})$ (red bars) and the average of $L_1$ norm of $\textbf{P}_i^{(j)}$ (blue bars) on each layer, according to Eq.~(\ref{eqn:gate}). The values are extracted for the lexical aspect (top left), temporal aspect (top right), and French synonymous sentences (bottom).
  }
  \label{fig:stability-mt}
\end{figure} 

For \textsc{GeDi}~\cite{krause-etal-2021-gedi-generative},
\textsc{Dist. Lens}~\cite{DBLP:journals/corr/abs-2210-02889}, and
\textsc{Tailor}~\cite{DBLP:journals/corr/abs-2204-13362}, we follow the settings in their paper. 
Specifically, we found that the weights for attribute balance and the number of candidates in the decoding stage of \textsc{Dist. Lens} significantly affect constraint accuracies. For the weights for attribute balance and the number of candidates, we searched in \{$0.1$, $0.2$, $0.5$, $1$, $1.5$, $2$, $3$, $4$, $5$, $8$, $10$, $25$, $50$\} and \{$1$, $2$, $5$, $10$\}, respectively. 
For the other methods, we demonstrate their hyper-parameters in Table~\ref{tab:common_hyper_para}. 
Table~\ref{tab:num_parameters} shows the number of trainable parameters of each method. 
The learning rates were determined by searching in \{$1\times10^{-5}$, $2\times10^{-5}$, $4\times10^{-5}$, $8\times10^{-5}$, $1\times10^{-4}$, $2\times10^{-4}$, $4\times10^{-4}$, $1\times10^{-3}$\} on the development set. In our 
 approach, the textual contexts used for attribute-based constraints (see \S\ref{sec:method}) are:
 \begin{itemize}
     \item Sentiment: ``This is a \{\} review.'' for ``positive/negative''.
     \item Topic: ``The following is about \{\} food.'' for ``Asian/American/Mexican''.
     \item Tense: ``The tense of this sentence is the \{\} tense.'' for ``past/present/future'', and ``The tense of this sentence is undecided.'' for sentences that do not have an explicit tense.
 \end{itemize}
Note that our use of existing artifact(s) was consistent with their intended use. The source code of this work is available at \url{https://github.com/THUNLP-MT/PromptGating4MCTG} and it was developed based on THUMT~\cite{tan-etal-2020-thumt}, an open-source toolkit for machine translation.

\begin{table}[!t]
    \begin{center}
    \scalebox{0.7}{
        \begin{tabular}{l|cc}
        \toprule
        \multirow{2}{*}{\bf Method}& \multicolumn{2}{c}{\bf{Speed}} \\
        &  {\em Training} (hours) & {\em Inference} (sec/sent.)  \\\midrule \midrule
        \textsc{Gedi} & \ \ 5.4032 & 1.2020 \\
        \textsc{Dist. Lens} & 30.1320 & 2.5705 \\
        {\textsc{Prompt-tuning}}& \ \ 4.0983 & 0.2122 \\
        {\textsc{Prefix-tuning}} & \ \ 3.9025 & 0.2220 \\
        \textsc{Tailor} & \ \ 4.1055 & 0.4640 \\
        {\textsc{Prompt Gating}}& \ \ 4.5204 & 0.2108 \\
        \bottomrule
        \end{tabular}
        }
    \caption{The training and inference speeds of each method on multi-aspect controllable text generation. The training speeds are presented as the average training time on a single aspect, and the inference speeds are displayed in the form of time per sentence. Note that the inference procedure of ``Dist. Lens'' includes training a K-Center model~\cite{DBLP:journals/corr/abs-2210-02889}.}
    \label{tab:efficiency}
    \end{center}
\end{table}

\begin{table}[!t]
    \begin{center}
    \scalebox{0.7}{
        \begin{tabular}{l|cc}
        \toprule
        \multirow{2}{*}{\bf Method}& \multicolumn{2}{c}{\bf{Speed}} \\
        &  {\em Training} (hours) & {\em Inference} (sec/sent.)  \\\midrule \midrule
        {\textsc{Prefix-tuning}} & 10.0150 & 0.2220 \\
        {\textsc{LoRA}}& 10.8805 & 0.1920 \\
        \textsc{Parallel Adapter}    & 12.0330 & 0.2150 \\
        {\textsc{Prompt-tuning}}&\ \ 8.7225 & 0.2122 \\
        {\textsc{Prompt Gating}}& 11.0330 & 0.2108 \\
        \bottomrule
        \end{tabular}
        }
    \caption{The training and inference speeds of each method on multi-aspect controllable machine translation. The training speeds are presented as the average training time on a single aspect, and the inference speeds are displayed in the form of time per sentence.}
    \label{tab:efficiency2}
    \end{center}
\end{table}

\section{Experimental Results} \label{sec:app_exp}

\subsection{More Analyses} \label{app:analyses}

\paragraph{Visualization of \textsc{Prompt Gating}.}

To further investigate how our approach alleviates the mutual interference of plugins, we visualized the trainable parameters in \textsc{Prompt Gating}. Specifically, we first extracted $\textbf{P}_i^{(j)}$ and $\sigma (\textbf{G}_i^{(j)})$ in Eq.~(\ref{eqn:gate}) from each layer $j$ for every single aspect. Then we calculated the average of $\sigma (\textbf{G}_i^{(j)})$ and the $L_1$ norm of $\textbf{P}_i^{(j)}$ over all the layers, which are represented by red and blue bars respectively in Figure~\ref{fig:stability} and Figure~\ref{fig:stability-mt}.

We can find that when the magnitude of $\textbf{P}_i^{(j)}$ (i.e., trainable vectors added to hidden states) becomes larger, the values of $\sigma (\textbf{G}_i^{(j)})$ (i.e., trainable gates) tend to become smaller. In other words, these trainable gates attempt to normalize or stabilize the magnitude of hidden states and thus alleviate mutual interference.

\paragraph{Efficiency.}

Table~\ref{tab:efficiency} and Table~\ref{tab:efficiency2} show the training and inference speeds of each method in text generation and machine translation. All
training and inference were run on a single GeForce RTX 3090 GPU.

\subsection{Detailed Results} \label{app:main_results}

Table~\ref{tab:app_ctg} and ~\ref{tab:app_mt} are the detailed versions of Table~\ref{tab:ctg} and~\ref{tab:mt}, respectively. 
We provide detailed results of both single- and multi-aspect models. For text generation, we further demonstrate the accuracy of each attribute.

\section{Case Study}  \label{app:case_study}

To further investigate the fulfillment and text quality of each combination of constraints of these methods, Table~\ref{tab:main1_example} and Table~\ref{tab:main2_example} demonstrate examples of  text generation and machine translation, respectively. Models only trained on single-aspect data are required to give results satisfying multiple aspects of constraints.

\section{Details in Human Evaluation}
In this section, we show more details about the human annotation adopted for evaluating model performance on text generation. We recruited three volunteers from schools, shuffled the output of models and provided it to them for scoring. Since they are volunteers, they were not paid. Their average age is 25 years old and they have enough daily English communication skills. The instruction we provided to them like ``This human evaluation aims to evaluate the model-generated review texts in three aspects: sentiment and topic relevance, and text fluency. All three integer scores are on a scale of 1-5, with a higher degree of topic/sentiment relevance representing a more consistent theme/sentiment, and a higher degree of text fluency representing a more fluent text. Your personal information will not be retained and these scores will only be used for human evaluation in research''.

\onecolumn
\begin{small}
\begin{longtable}{l|l|rr|rrr|r}
            \toprule
            \multirow{2}{*}{\bf Method} & \multirow{2}{*}{\bf Constraint} & \multicolumn{2}{c|}{\bf Sentiment$\uparrow$} &  \multicolumn{3}{c|}{\bf Topic$\uparrow$}  & \multirow{2}{*}{\bf PPL$\downarrow$} \\
            
            
            & & \em Positive & \em Negative & \em Asian & \em American & \em Mexican  & \\
            \midrule \midrule

\multicolumn{8}{c}{\em Decoding-Time Regulation Method} \\\midrule
            
\multirow{13}{*}{\textsc{Gedi}} & \multirow{5}{*}{\em single-aspect} & 98.67 & /      & /     & /     & /     & 227.87  \\
                         &                                & /     & 100.00 & /     & /     & /     & 839.69  \\
                         &                                & /     & /      & 94.93 & /     & /     & 206.97  \\
                         &                                & /     & /      & /     & 99.73 & /     & 246.36  \\
                         &                                & /     & /      & /     & /     & 97.33 & 297.54  \\
      \cmidrule{2-8}     & \multirow{6}{*}{\em multi-aspect}  & 98.67 & /      & 28.27 & /     & /     & 363.51  \\
                         &                                & 99.20 & /      & /     & 87.73 & /     & 1834.65 \\
                         &                                & 99.47 & /      & /     & /     & 37.87 & 378.73  \\
                         &                                & /     & 100.00 & 44.53 & /     & /     & 329.38  \\
                         &                                & /     & 99.73  & /     & 97.87 & /     & 423.21  \\
                         &                                & /     & 99.73  & /     & /     & 11.87 & 372.03  \\
      \cmidrule{2-8}     & \em single (avg.)                  & 98.67 & 100.00 & 94.93 & 99.73 & 97.33 & 363.69  \\
      \cmidrule{2-8}     & \em multi (avg.)                   & 99.11 & 99.82  & 36.40 & 92.80 & 24.87 & 616.92  \\
                         \midrule

\multicolumn{8}{c}{\em Parameter Efficient Tuning Method with Joint Training} \\\midrule

\multirow{13}{*}{\textsc{Dist. Lens}} & \multirow{5}{*}{\em single-aspect} & 91.33 & /     & /     & /     & /     & 28.48   \\
                         &                                & /     & 97.95 & /     & /     & /     & 28.70   \\
                         &                                & /     & /     & 77.33 & /     & /     & 39.01  \\
                         &                                & /     & /     & /     & 88.98 & /     & 29.87   \\
                         &                                & /     & /     & /     & /     & 79.47 & 38.26   \\
      \cmidrule{2-8}     & \multirow{6}{*}{\em multi-aspect}  & 36.27 & /     & 43.73 & /     & /     & 45.89 \\
                         &                                & 57.87 & /     & /     & 71.73 & /     & 49.84 \\
                         &                                & 74.67 & /     & /     & /     & 54.67 & 47.59 \\
                         &                                & /     & 99.73 & 70.13 & /     & /     & 56.20 \\
                         &                                & /     & 97.60 & /     & 78.93 & /     & 59.24 \\
                         &                                & /     & 98.67 & /     & /     & 82.67 & 56.77 \\
      \cmidrule{2-8}     & \em single (avg.)                  & 91.33 & 97.95 & 77.33 & 88.98 & 79.47 & 32.86   \\
      \cmidrule{2-8}     & \em multi (avg.)                   & 56.27 & 98.67 & 56.93 & 75.33 & 68.67 & 52.59 \\
                         \midrule

\multicolumn{8}{c}{\em Parameter Efficient Tuning Methods without Joint Training} \\\midrule

\multirow{13}{*}{\textsc{Prompt-tuning}} & \multirow{5}{*}{\em single-aspect} & 52.00 & /     & /     & /     & /     & 136.10 \\
                         &                                & /     & 56.27 & /     & /     & /     & 26.26  \\
                         &                                & /     & /     & 46.67 & /     & /     & 25.83  \\
                         &                                & /     & /     & /     & 84.53 & /     & 26.07  \\
                         &                                & /     & /     & /     & /     & 39.60 & 24.36  \\
      \cmidrule{2-8}     & \multirow{6}{*}{\em multi-aspect}  & 57.07 & /     & 40.80 & /     & /     & 65.96  \\
                         &                                & 59.47 & /     & /     & 83.87 & /     & 30.45  \\
                         &                                & 45.87 & /     & /     & /     & 20.13 & 47.43  \\
                         &                                & /     & 49.73 & 30.27 & /     & /     & 40.78  \\
                         &                                & /     & 38.67 & /     & 84.00 & /     & 33.43  \\
                         &                                & /     & 38.93 & /     & /     & 29.60 & 27.30  \\
      \cmidrule{2-8}     & \em single (avg.)                  & 52.00 & 56.27 & 46.67 & 84.53 & 39.60 & 47.72  \\
      \cmidrule{2-8}     & \em multi (avg.)                   & 54.13 & 42.44 & 35.53 & 83.93 & 24.87 & 40.89  \\
                         \midrule

\multirow{13}{*}{\textsc{Prefix-tuning}} & \multirow{5}{*}{\em single-aspect} & 70.67 & /     & /     & /     & /     & 23.53  \\
                         &                                & /     & 98.93 & /     & /     & /     & 21.55  \\
                         &                                & /     & /     & 77.87 & /     & /     & 21.89  \\
                         &                                & /     & /     & /     & 84.00 & /     & 21.99  \\
                         &                                & /     & /     & /     & /     & 73.60 & 22.55  \\
      \cmidrule{2-8}     & \multirow{6}{*}{\em multi-aspect}  & 64.53 & /     & 70.27 & /     & /     & 141.71 \\
                         &                                & 67.20 & /     & /     & 80.00 & /     & 243.93 \\
                         &                                & 51.20 & /     & /     & /     & 65.73 & 125.30 \\
                         &                                & /     & 33.87 & 60.67 & /     & /     & 118.65 \\
                         &                                & /     & 27.32 & /     & 78.13 & /     & 138.91 \\
                         &                                & /     & 41.07 & /     & /     & 59.87 & 116.34 \\
      \cmidrule{2-8}     & \em single (avg.)                  & 70.67 & 98.93 & 77.87 & 84.00 & 73.60 & 22.30  \\
      \cmidrule{2-8}     & \em multi (avg.)                   & 60.98 & 34.08 & 65.47 & 79.07 & 62.80 & 147.47 \\
\midrule
\multirow{2}{*}{\bf Method} & \multirow{2}{*}{\bf Constraint} & \multicolumn{2}{c|}{\bf Sentiment$\uparrow$} &  \multicolumn{3}{c|}{\bf Topic$\uparrow$}  & \multirow{2}{*}{\bf PPL$\downarrow$} \\

& & \em Positive & \em Negative & \em Asian & \em American & \em Mexican  & \\
\midrule \midrule
\multirow{13}{*}{\textsc{Tailor}} & \multirow{5}{*}{\em single-aspect} & 81.87 & /     & /     & /     & /     & 32.80 \\
                         &                                & /     & 95.73 & /     & /     & /     & 24.21 \\
                         &                                & /     & /     & 72.00 & /     & /     & 34.38 \\
                         &                                & /     & /     & /     & 88.00 & /     & 33.35 \\
                         &                                & /     & /     & /     & /     & 76.00 & 34.10 \\
      \cmidrule{2-8}     & \multirow{6}{*}{\em multi-aspect}  & 72.73 & /     & 67.47 & /     & /     & 43.08 \\
                         &                                & 72.53 & /     & /     & 70.07 & /     & 32.87 \\
                         &                                & 68.27 & /     & /     & /     & 69.33 & 43.12 \\
                         &                                & /     & 90.67 & 68.00 & /     & /     & 44.64 \\
                         &                                & /     & 90.40 & /     & 70.93 & /     & 33.54 \\
                         &                                & /     & 89.47 & /     & /     & 66.53 & 44.50 \\
      \cmidrule{2-8}     & \em single (avg.)                  & 81.87 & 95.73 & 72.00 & 88.00 & 76.00 & 31.77 \\
      \cmidrule{2-8}     & \em multi (avg.)                   & 71.18 & 90.18 & 67.73 & 70.50 & 67.93 & 40.29 \\
                         \midrule

\multirow{13}{*}{\textsc{Prompt Gating} (\textit{Ours})} & \multirow{5}{*}{\em single-aspect} & 91.73 & /     & /     & /     & /     & 21.65 \\
                         &                                & /     & 99.73 & /     & /     & /     & 20.39 \\
                         &                                & /     & /     & 77.87 & /     & /     & 21.29 \\
                         &                                & /     & /     & /     & 89.87 & /     & 21.88 \\
                         &                                & /     & /     & /     & /     & 81.33 & 22.93 \\
      \cmidrule{2-8}     & \multirow{6}{*}{\em multi-aspect}  & 73.07 & /     & 62.13 & /     & /     & 21.28 \\
                         &                                & 77.60 & /     & /     & 82.93 & /     & 21.65 \\
                         &                                & 75.20 & /     & /     & /     & 72.00 & 22.47 \\
                         &                                & /     & 93.33 & 73.87 & /     & /     & 21.64 \\
                         &                                & /     & 95.73 & /     & 81.33 & /     & 20.09 \\
                         &                                & /     & 93.87 & /     & /     & 77.87 & 23.50 \\
      \cmidrule{2-8}     & \em single (avg.)                  & 91.73 & 99.73 & 77.87 & 89.87 & 81.33 & 21.63 \\
      \cmidrule{2-8}     & \em multi (avg.)                   & 75.29 & 94.31 & 68.00 & 82.13 & 74.93 & 21.77 \\
      
            \bottomrule
        \caption{\label{tab:app_ctg} Detailed results of automatic evaluation on double-aspect controllable text generation. ``{\em single (avg.)}'' denotes the average score over scores in the single-aspect setting. ``{\em multi (avg.)}'' denotes the average score over scores in the multi-aspect setting.
        } 
\end{longtable}
\end{small}
\twocolumn

\begin{table*}
    \centering
    \small
    \begin{center}
        \begin{tabular}{l|l|rr|r}
            \toprule
            \bf Method & \bf Constraint & \bf Lex. $\uparrow$ & \bf Tense$\uparrow$ & \bf BLEU$\uparrow$ \\
            \midrule \midrule
            
\multicolumn{2}{c|}{\em w/o control}                 & 50.98 & 79.52 & 32.7 \\\midrule
\multirow{5}{*}{\textsc{Prefix-tuning}}    & \em lexical   & 85.28 & 80.99 & 36.7 \\
                                  & \em temporal             & 50.80 & 83.28 & 33.0 \\
                                  & \em knowledgeable & 50.20 & 79.29 & 32.8 \\\cmidrule{2-5}
                                  & \em single (max.)  & 85.28 & 83.28 & 36.7 \\\cmidrule{2-5}
                                  & \em multi (avg.)     & 7.51  & 43.46 & 0.4  \\\midrule
\multirow{5}{*}{\textsc{Parallel Adapter}} & \em lexical   & 91.82 & 81.12 & 37.3 \\
                                  & \em temporal             & 50.93 & 83.55 & 33.1 \\
                                  & \em knowledgeable & 50.16 & 79.35 & 32.8 \\\cmidrule{2-5}
                                  & \em single (max.)  & 91.82 & 83.55 & 37.3 \\\cmidrule{2-5}
                                  & \em multi (avg.)     & 48.44 & 67.87 & 21.8 \\\midrule
\multirow{5}{*}{\textsc{LoRA}}             & \em lexical   & 87.94 & 81.59 & 36.2 \\
                                  & \em temporal             & 50.79 & 84.32 & 33.0 \\
                                  & \em knowledgeable & 50.57 & 80.22 & 32.7 \\\cmidrule{2-5}
                                  & \em single (max.)  & 87.94 & 84.32 & 36.2 \\\cmidrule{2-5}
                                  & \em multi (avg.)     & 50.79 & 74.16 & 25.0 \\\midrule
\multirow{5}{*}{\textsc{Prompt-tuning}}    & \em lexical   & 74.93 & 80.99 & 35.4 \\
                                  & \em temporal             & 50.25 & 81.19 & 33.0 \\
                                  & \em knowledgeable & 50.17 & 78.82 & 32.5 \\\cmidrule{2-5}
                                  & \em single (max.)  & 74.93 & 81.19 & 35.4 \\\cmidrule{2-5}
                                  & \em multi (avg.)     & 64.64 & 81.12 & 34.2 \\\midrule
\multirow{5}{*}{\textsc{Prompt Gating} (\textit{Ours})}    & \em lexical   & 89.90 & 81.29 & 37.1 \\
                                  & \em temporal                & 50.95 & 83.98 & 33.2 \\
                                  & \em knowledgeable & 51.18 & 79.55 & 33.1 \\\cmidrule{2-5}
                                  & \em single (max.)  & 89.90 & 83.98 & 37.1 \\\cmidrule{2-5}
                                  & \em multi (avg.)     & 85.29 & 85.75 & 36.8 \\
                                  
            \bottomrule
        \end{tabular}
        \caption{\label{tab:app_mt} Detailed results on controllable machine translation. The experiments are conducted on the WMT14 German$\rightarrow$English benchmark. There are three aspects of constraints: lexical constraint, tense, and external knowledge (French synonymous sentence). ``{\em w/o control}'': the pretrained German$\rightarrow$English Transformer translation model without any plugin. ``{\em single (max.)}'' denotes the maximal score among scores in the single-aspect setting. ``{\em multi (avg.)}'' denotes the average score over scores in the multi-aspect setting.
        }
    \end{center}
\end{table*}

\begin{table*}[!ht]
    \begin{center}
{\large\textcolor{red}{WARNING: Next may contain contents that are offensive in nature.}}
        \resizebox{\linewidth}{!}{
        \begin{tabular}{@{}c|l|l@{}}
\toprule
\textbf{Constraints}                                   & \multicolumn{1}{c|}{\textbf{Method}}           & \multicolumn{1}{c}{\textbf{Generation Result}}       \\ \midrule

{\color[HTML]{FF0000} }                               &                                                 & \underline{While this is happening} to people of colour over and over again in schools and hospitals regardless of             \\
&                                                                                                       & how kids should behave, let alone thats happening here? Is we as a society humane towards \\
{\color[HTML]{FF0000} }                               & \multirow{-3}{*}{\textsc{GeDi}}                 &  minks that poo on each other before death leaves them.    \\ \cmidrule(l){2-3} 
&                                                                                                       & \underline{While this   is happening} i was very impressed. food is average and the service is  \textcolor{red}{great}. \\
{\color[HTML]{FF0000} }                               & \multirow{-2}{*}{\textsc{Dist. Lens}}           &    Advertisements: -)   ***EDIT: The   vegetarian menu is  \textcolor{red}{amazing}!                                                                                 \\ \cmidrule(l){2-3} 
{\color[HTML]{FF0000} }                               &                                                 & \underline{While this is happening} in my hometown. In  \textcolor{red}{New York, New York, and the UK} , are in need of something,                                                                                                                               \\
{\color[HTML]{FF0000} }                               &                                                 & they say. It's a good year in my blood. In  \textcolor{red}{New Orleans}, is it is good for the heart to get rid of              \\
&                                                                                                       & it.In  \textcolor{red}{New York} is getting all of its emotions out because we don't want to make the traditional  church    \\
{\color[HTML]{FF0000} }                               & \multirow{-4}{*}{\textsc{Prompt-tuning}}        & in-state.It was also all the land that is, and it is \textcolor{blue}{not good} is all are for me is all a   rock, and we don’                                                                         \\ \cmidrule(l){2-3} 
{\color[HTML]{FF0000} }                               & \textsc{Prefix-tuning}                          & \underline{While this   is happening} food. is about  \textcolor{darkgreen}{Asian}.    \textcolor{red}{isaboutAsian.is-foodsfood-trending.s.e...trend-stings-food-trends }                                                                                                    \\ \cmidrule(l){2-3} 
{\color[HTML]{FF0000} }                               &                                                 & \underline{While this is happening} in my hometown, i will say that it's a little  \textcolor{blue}{pricey} but worth it for a  \textcolor{red}{good} meal.                                                                                        \\
&                                                                                                       & the food was  \textcolor{red}{good}, the service was \textcolor{red}{good}, the service was \textcolor{red}{fast} and the food was \textcolor{red}{tasty}. the food  \\ 
{\color[HTML]{FF0000} }                               &                                                 & \textcolor{red}{arrived on time},  so i could eat it. i did get a couple of things \textcolor{blue}{wrong with this place}.  \textcolor{darkgreen}{the rice} is \textcolor{blue}{ bland}.        \\
{\color[HTML]{FF0000} }                               &                                                 & i think \textcolor{darkgreen}{the rice} is \textcolor{blue}{too salty}. my boyfriend ordered the fried squid. it was \textcolor{blue}{a bit greasy}.  the shrimp had                                                                                                                              \\
{\color[HTML]{FF0000} }                               & \multirow{-5}{*}{\textsc{Tailor}}               &  \textcolor{red}{a good flavor}  and it was   \textcolor{red}{fresh}. i \textcolor{red}{would give it a try again} but it won't be here for at least 3 years.                                                                                                                     \\ \cmidrule(l){2-3} 
{\color[HTML]{FF0000} }                               &                                                 & \underline{While this is happening}, i've never been here to the \textcolor{darkgreen}{japanese section} and have never                                             \\
&                                                                                                       &  eaten here before. the place is \textcolor{blue}{small} and the place is \textcolor{blue}{crowded}. the food is \textcolor{blue}{small}, and the service   \\
{\color[HTML]{FF0000} }                               &                                                 &  \textcolor{blue}{isn't very friendly}. i did not expect to spend a lot of money for \textcolor{darkgreen}{a sushi place} because it's pretty small.    \\
&                                                                                                       & they have a few places where they can take a large order and they give it to those  who pay \\
\multirow{-20}{*}{\color{blue}{Neg.}, \textcolor{darkgreen} {Asian}} & \multirow{-5}{*}{\textsc{Prompt Gating} (\textit{Ours})} &      for the food. also the staff \textcolor{blue}{seems to ignore you} if you come for \textcolor{darkgreen}{sushi}. \\ \midrule
&                                                                                                       & \underline{The food} \textcolor{blue}{sucks!!} \textcolor{red}{Luckily} we're   Gilmore Girls and I still \textcolor{red}{feel pretty strong} \\
& \multirow{-2}{*}{\textsc{GeDi} }                                                                      & from an adrenaline rush, not   being adequately clothed.  Return to   Content                                                                                 \\ \cmidrule(l){2-3} 
&                                                                                                       &  \underline{The food} is \textcolor{red}{great} and  \textcolor{blue}{a little   overpriced}. service was \textcolor{red}{great though}. i will \textcolor{blue}{ never go back.} \\
   & \multirow{-2}{*}{\textsc{Dist. Lens}}                                                           &  Advertisements   for the  \textcolor{red}{sake of getting good food}                                                                                   \\ \cmidrule(l){2-3} 
&                                                                                                       & \underline{The food} you don't need to worry about it. This is the  \textcolor{red}{most important food}. This could be years.                                                                                \\
&                                                                                                   & This would also have you look for new business to move to. This is a big one \\
&                                                                                                   & and we're not   looking for one.This might get more people. It may get people \\
  & \multirow{-4}{*}{\textsc{Prompt-tuning}}                                                        & looking for   work elsewhere, it could get people  \textcolor{red}{lookin} at work in the area.                                                               \\ \cmidrule(l){2-3} 
 &                                                                                                  & \underline{The food}.  \textcolor{blue}{is about negative aspects}, and    \textcolor{blue}{about negative food}.” – Michael Siegel. \\
 & \multirow{-2}{*}{\textsc{Prefix-tuning}}                                                         &  following is  \textcolor{blue}{a negative review}?   –@siegel.                                                                                                              \\ \cmidrule(l){2-3} 
  &                                                                                                 & \underline{The food} was \textcolor{red}{good} for what you paid for it, but the service was \textcolor{blue}{terrible}, the food \textcolor{blue}{tasted bad}.                         \\
&                                                                                                   &  the waiter was \textcolor{blue}{rude} and was constantly \textcolor{blue}{taking orders that were not on the table}.  \\
&                                                                                                   & when he asked why the food was so hot or that they were out of food.                               \\
&                                                                                                   &  he said that the menu was out of order because of a busy night. after that he told me to call him    \\
&                                                                                                   & back to make sure that the order was in order. the food itself was \textcolor{blue}{not that great},   \\
& \multirow{-6}{*}{\textsc{Tailor}}                                                                   &    but \textcolor{red}{that's ok} because it was only \$6. \\ \cmidrule(l){2-3} 
&                                                                                                     & \underline{The food} was \textcolor{blue}{very bland}. i had a \textcolor{darkgreen}{burger}   that was \textcolor{blue}{really dry}. they had the \textcolor{darkgreen}{"tender beef patty"}. \\
\multirow{-18}{*}{ \textcolor{blue}{Neg.},  \textcolor{darkgreen}{American}} & \multirow{-2}{*}{\textsc{Prompt Gating} (\textit{Ours})} &  \textcolor{darkgreen}{the beef} was \textcolor{blue}{dry} so i \textcolor{blue}{didn't like} the "tasty" flavor and \textcolor{darkgreen}{the burger} \textcolor{blue}{tasted stale}. i \textcolor{blue}{don't eat that type of beef}.                       \\ \midrule
& \textsc{GeDi}                                                     & \underline{The country} \textcolor{blue}{didn't learn its lesson} soon.                                                                                                                                                                                                   \\ \cmidrule(l){2-3} 
&                                                                   & \underline{The country} \textcolor{red}{fried is great}, \textcolor{blue}{not the best}.   \textcolor{red}{service was decent and we'll be back!} \\
& \multirow{-2}{*}{\textsc{Dist. Lens}}                             &  Advertisements are \textcolor{red}{good!}   Related Posts: (                                                                                                               \\ \cmidrule(l){2-3} 
&                                                                   & \underline{The country}. hell yes I am. and I am off. I am off to do some drugs and drink with a woman. I am off.      \\
&                                                                   & and I am also off on drugs with the women! I can see I could go. and have sex with someone.  \\
&                                                                   & I don't think I am a homa, though I could have sex sometime,  which \textcolor{blue}{wouldn't be good with} the men.I  \\
  & \multirow{-4}{*}{\textsc{Prompt-tuning}}                        &  am off my period and I'm not ready yet   for that, except sex, it is \textcolor{red}{good with} the women                                                 \\ \cmidrule(l){2-3} 
  &                                                                 & \underline{The country} and \textcolor{darkgreen}{Mexican food}. is about \textcolor{darkgreen}{Mexican food}. for a reason. It's about \textcolor{darkgreen}{Mexican food}.                                                                                \\
&                                                                   & following is about \textcolor{red}{Chinese food}.is...Read Full Article......and more »Zachary: \textcolor{red}{Food is delicious...!}\}\textgreater{} \\      
   & \multirow{-3}{*}{\textsc{Prefix-tuning}}                       & More...\textless{}/show\textgreater{}"\textgreater{}Chef-in-chief''...Zach!\textless{}/show\textgreater{}\textless{}Chev..."\textgreater{}ZACH!....!"\textless{}/show\textgreater{} \\ \cmidrule(l){2-3} 
    &                                                               & \underline{The country} style, \textcolor{red}{fresh, delicious} \textcolor{darkgreen}{tacos}, and \textcolor{red}{fresh} \textcolor{darkgreen}{tortilla chips}. \textcolor{darkgreen}{the salsa, the burrito} and  \\
    &                                                               & \textcolor{red}{rice were good}. \textcolor{darkgreen}{the salsa} was \textcolor{red}{very good}, but the rice was \textcolor{blue}{a bit dry}. i'm going to \textcolor{red}{give the burrito a 10/10}.      \\
&                                                                   &  \textcolor{darkgreen}{the burrito} was \textcolor{blue}{not as fresh as i would like}, and \textcolor{darkgreen}{the salsa} was \textcolor{blue}{lacking, not fresh enough}. my brother got   \\
  & \multirow{-4}{*}{\textsc{Tailor}}                               & \textcolor{darkgreen}{the burrito with the green salsa}, he \textcolor{red}{liked it}.                        \\ \cmidrule(l){2-3} 
 &                                                                  & \underline{The country} food at this place has become \textcolor{blue}{so bland that i have stopped coming here}.                  \\
&                                                                   & \textcolor{blue}{the service was slow and the food didn't have any flavor to it at all}. it just \textcolor{blue}{felt stale} to the touch.  \\
&                                                                    & i have never been in the place. i have been to many different \textcolor{darkgreen}{mexian restaurants} in   my time in vegas,                                                                          \\ 
\multirow{-18}{*}{\textcolor{blue}{Neg.}, \textcolor{darkgreen}{Mexican}} & \multirow{-3}{*}{\textsc{Prompt Gating} (\textit{Ours})} &  however i can only say they are \textcolor{blue}{the worst in vegas}, at   least to this point, in my memory \\ \midrule
\end{tabular}}
    \end{center}
\end{table*}
\begin{table*}[!ht]
\begin{center}
\resizebox{\linewidth}{!}{
\begin{tabular}{@{}c|l|l@{}}
\toprule
\textbf{Constraints}                                   & \multicolumn{1}{c|}{\textbf{Method}}           & \multicolumn{1}{c}{\textbf{Generation Result}} \\ \midrule
\multicolumn{1}{l|}{}                                 &  \textsc{GeDi}                                              & \makecell[l]{\underline{Once upon a time}…freedom—an old story…"" As Vegeta's hands encased his claws, Virankhan \\ extended them to enhance their effects. His powers - including  Frieza's tail transformation, long range \\ ki blasts and history-defining transformations  like Ash Ketchum's samurai thrill and Bulma \& \\ Pikkonü putting on their samurai mastership   – swept through the crowd  as they \textcolor{blue}{cheerfully flooded} \\ gutter comforts grounds majors like dubbed   Kingston area.   His qir } \\ \cmidrule(l){2-3} 
\multicolumn{1}{l|}{} & \textsc{Dist. Lens} & \makecell[l]{\underline{Once upon a time}, \textcolor{blue}{great food}. the service   is \textcolor{blue}{great and very friendly}. \\ Advertisements - \textcolor{blue}{yummy!}   *** Helpful!! :)}   \\ \cmidrule(l){2-3} 
\multicolumn{1}{l|}{}  &  \textsc{Prompt-tuning}     & \makecell[l]{\underline{Once upon a time} travel to another part of the park and back to this point, the park was still full of black. \\And this was the point after which we were left back there.   And now I want to get up and go for another part. \\ This is going back to this point but we are still left back in town to get rid of this.And you're left with that  \\ spot  in the park.And you're left at the end of time for two. I'm getting rid from it but it has to go back} \\ \cmidrule(l){2-3} 
\multicolumn{1}{l|}{}                                 & \textsc{Prefix-tuning}                          & \makecell[l]{\underline{Once upon a time}. following is about   food.Advertisements…comments…comments}    \\ \cmidrule(l){2-3} 
\multicolumn{1}{l|}{}    &   \textsc{Tailor}  & \makecell[l]{\underline{Once upon a time}, i was \textcolor{blue}{a big fan of} this establishment. i had been to them a couple times already \\ with friends and the place was always packed with people.    this place was \textcolor{blue}{a hit},\\ so much so that now I am going to go there every time.}    \\ \cmidrule(l){2-3} 
\multicolumn{1}{l|}{\multirow{-18}{*}{\textcolor{blue}{Pos.}, \textcolor{darkgreen}{Asian}}}   &   \textsc{Prompt Gating} (\textit{Ours})    & \makecell[l]{\underline{Once upon a time} i was at the restaurant and had dinner at a friend's place, the food was \textcolor{blue}{great}. \\ the chef is \textcolor{blue}{very knowledgeable} (the owner's daughter is \textcolor{darkgreen}{a sushi professional}), but i can only \\ say that the restaurant was \textcolor{blue}{very clean} and the food was \textcolor{blue}{very flavorful}.}  \\ \midrule

\multicolumn{1}{l|}{}     &     \textsc{GeDi}    & \makecell[l]{\underline{The year is 1910.} Ephraim Legrand spurges explores Europe's forgotten North Atlantic island, exploring \\
its \textcolor{blue}{rich history, diverse landscapes, richside civic pride and communion}.     Through photography held in private \\ property   events concentrated throughout the year, content creators can collaborate   creatively and celebrate \\ individuality night-out!}    \\ \cmidrule(l){2-3} 
\multicolumn{1}{l|}{}    & \textsc{Dist. Lens}       & \makecell[l]{\underline{The year is 1910.} \textcolor{blue}{great food and the service was great}. \textcolor{blue}{will be back!} \\ Advertisements for \textcolor{red}{good sushi}!!! Follow me   @wendie  }\\ \cmidrule(l){2-3} 
\multicolumn{1}{l|}{}         & \textsc{Prompt-tuning}    & \makecell[l]{\underline{The year is 1910.} We are still going out to dinner. This time we're \textcolor{blue}{having fun} with it but I don't like \\to leave it for another night.And this night has just been getting started  so it's time for everyone \\ to leave.And we've been eating this night for a week but \textcolor{blue}{not yet ready to leave the place} to go to bed \\ yet.And we've just eaten.But this morning has just     started leaving this place\\ so I am getting over it.You're   not over it yet."And yet he is  }   \\ \cmidrule(l){2-3} 
\multicolumn{1}{l|}{}     & \textsc{Prefix-tuning}                          & \makecell[l]{\underline{The year is 1910.} is about \textcolor{darkgreen}{American food}.   is…\\http://news-and-review-food-tourism.org/article.asp/?articleId=9c0c4-a2a-e0-b9-a-9c5d7a8b8f0\_0\_0.0   }        \\ \cmidrule(l){2-3} 
\multicolumn{1}{l|}{} & \textsc{Tailor}      & \makecell[l]{\underline{The year is 1910.} this is not your average \textcolor{red}{sushi} \textcolor{darkgreen}{bar}. the atmosphere was \textcolor{blue}{very casual and cozy}. \\the food was   \textcolor{blue}{good}. the service was \textcolor{blue}{prompt and fast}. the food was \textcolor{blue}{fresh and delicious}.}    \\ \cmidrule(l){2-3}
\multicolumn{1}{l|}{\multirow{-17}{*}{\textcolor{blue}{Pos.}, \textcolor{darkgreen}{American}}} & \textsc{Prompt Gating} (\textit{Ours}) & \makecell[l]{\underline{The year is 1910.} we went to this place for the first year. it \textcolor{blue}{looks nice}. there are a couple of booths, a table \\for 4 (and a counter) to get seated in. we had an early reservation. i ordered \textcolor{darkgreen}{the steak}. \textcolor{blue}{the steak was great}. \\ my boyfriend \textcolor{blue}{enjoyed} \textcolor{darkgreen}{the chicken wings}. i was \textcolor{blue}{very impressed} by \textcolor{darkgreen}{the steak}. }    \\ \midrule
\multicolumn{1}{l|}{}   & \textsc{GeDi}  & \makecell[l]{\underline{While this is happening} in \textcolor{red}{Scotland}, though, we come a good distance north to Saltire Mountains National Park \\ where we sit beside mountains on the Welsh coast called Myrta Tor.   It's \textcolor{blue}{amazing to discover} more about these \\ \textcolor{blue}{beloved mountain peaks} that haven't seen much mainstream attention until now. We'll travel along \\ \textcolor{blue}{beautiful clifftees} via Priordeuorth during walking  \& hiking treks around Myrta Tor 'check out our full itinerary \\ here: Also   visit Stonyrithenview Chantry  } \\ \cmidrule(l){2-3} 
\multicolumn{1}{l|}{}    & \textsc{Dist. Lens}     & \makecell[l]{\underline{While this is happening}, \textcolor{blue}{great food}! \textcolor{darkgreen}{the margaritas} are \textcolor{blue}{fantastic} and \textcolor{blue}{friendly staff}.   Advertisements = \textcolor{blue}{good} \\ \textcolor{darkgreen}{mexican restaurant}!   Related } \\ \cmidrule(l){2-3} 
\multicolumn{1}{l|}{}    &   \textsc{Prompt-tuning}  & \makecell[l]{\underline{While this is happening}. It has already been happening. It's not like it's \textcolor{blue}{not a bad thing} to have it but \textcolor{red}{it's still bad}. \\ I \textcolor{red}{don't like} that either. It's got to get out of here before we can make  any significant changes to it this is \\ not that big thing anymore, but you don't have to do that. That's just a few days ago \textcolor{red}{in the USA}.This is the best \\ way I could have done it and not done yet but  this is \textcolor{red}{way worse than that}. It   has to  }  \\ \cmidrule(l){2-3} 
\multicolumn{1}{l|}{}                                 & \textsc{Prefix-tuning}  & \makecell[l]{\underline{While this is happening}!Advertisements?Sponsors?Media?  } \\ \cmidrule(l){2-3} 
\multicolumn{1}{l|}{}  &   \textsc{Tailor}     & \makecell[l]{\underline{While this is happening} to be \textcolor{blue}{my new favorite restaurant}, and \textcolor{blue}{my new favorite location}, i can't say that i've ever \\ been to this place with all the other reviews. there is \textcolor{blue}{a lot of good food} at the table,  but the wait staff \\ and food is just \textcolor{blue}{as great}! \textcolor{blue}{i love that} there   are \textcolor{red}{no waiters to take my orders}, but that doesn't mean the food \\ is bad. the \textcolor{blue}{food here was good}.}   \\ \cmidrule(l){2-3} 
\multicolumn{1}{l|}{\multirow{-19}{*}{\textcolor{blue}{Pos.}, \textcolor{darkgreen}{Mexican}}}    &  \textsc{Prompt Gating} (\textit{Ours})  & \makecell[l]{\underline{While this is happening} i'm going here for dinner for the first time. the food here was \textcolor{blue}{very, very good and very tasty!!} \\we ordered \textcolor{darkgreen}{a couple of different salads and some tacos}. i got \textcolor{darkgreen}{a vego beef taco}   with a spicy sauce (it is \textcolor{blue}{very good}). \\ i also got an onion rings (it does not have any onions, nor are there many onions in this recipe), and it was delicious!}    \\ \bottomrule
\end{tabular}}

        \caption{\label{tab:main1_example} Examples of multi-aspect controllable text generation. The given textual prefixes (see \S\ref{app:eval_metric}) are underlined. ``\textbf{Constraints}'' denotes the combination of constraints in sentiment and topic aspects. Some generation contents that {\color{blue} in consist with the sentimental constraint} are highlighted in blue, some generation contents that {\color{darkgreen} in consist with the topical constraint} are highlighted in green, and some generation contents that {\color{red} fail to satisfy constraints} are highlighted in red.}
    \end{center}
\end{table*}

\begin{table*}[!ht]
    \begin{center}
{\large\textcolor{red}{WARNING: Next may contain contents that are offensive in nature.}}
        \resizebox{\linewidth}{!}{
\begin{tabular}{@{}ll@{}}
\toprule
\multicolumn{2}{c}{\textit{Example 1}} \\ \midrule
\multicolumn{2}{c}{\textbf{Constraint}} \\ \midrule
\multicolumn{1}{l|}{\textbf{Keywords}}          & "\textcolor{blue}{This}", "\textcolor{blue}{bedroom}", "\textcolor{blue}{completely}"                                    \\ \midrule
\multicolumn{1}{l|}{\textbf{Tense}}                & The tense of this sentence is the \textcolor{blue}{past} tense.            \\ \midrule
\multicolumn{1}{l|}{\textbf{Knowledge (French)}}        & Cette chambre et une autre ont été   complètement brûlées. \\ \midrule 

\multicolumn{2}{c}{\textbf{Source and Reference}} \\ \midrule

\multicolumn{1}{l|}{\textbf{Source (German)}}        & Dieses und ein   weiteres Zimmer brannten vollständig aus. \\ 
\midrule
\multicolumn{1}{l|}{\textbf{Reference}}            & \textcolor{blue}{This} and another \textcolor{blue}{bedroom} \textcolor{blue}{were} \textcolor{blue}{completely}   burnt out.      \\ \midrule
\multicolumn{2}{c}{\textbf{Translation}}           \\ \midrule
\multicolumn{1}{l|}{\textsc{Prompt-tuning}}        & \textcolor{blue}{This} and another \textcolor{red}{room} \textcolor{blue}{burned} \textcolor{blue}{completely}.                   \\ \midrule
\multicolumn{1}{l|}{\textsc{Prefix-tuning}}        & \textcolor{blue}{This} \textcolor{red}{is} the \textcolor{red}{'room under the sun'}.                          \\ \midrule
\multicolumn{1}{l|}{\textsc{LoRA}}                 & \textcolor{blue}{This} and another \textcolor{red}{room} \textcolor{blue}{burned} out   \textcolor{blue}{completely}.             \\ \midrule
\multicolumn{1}{l|}{\textsc{Parallel Adapter}}     & \textcolor{blue}{This} \textcolor{red}{tense} and another \textcolor{red}{room} \textcolor{red}{is} \textcolor{blue}{completely}   burnt out.     \\ \midrule
\multicolumn{1}{l|}{\textsc{Prompt Gating} (\textit{Ours})} & \textcolor{blue}{This} and another \textcolor{blue}{bedroom} \textcolor{blue}{burned}   \textcolor{blue}{completely} out.          \\ \bottomrule
\toprule
\multicolumn{2}{c}{\textit{Example 2}} \\ \midrule
\multicolumn{2}{c}{\textbf{Constraint}} \\ \midrule
\multicolumn{1}{l|}{\textbf{Keywords}}                           & "\textcolor{blue}{The}", "\textcolor{blue}{transgender}", "\textcolor{blue}{employment,}"                                                                                                      \\ \midrule
\multicolumn{1}{l|}{\textbf{Tense}}                                 & The tense of this sentence is the \textcolor{blue}{present} tense.                                                                               \\ \midrule
\multicolumn{1}{l|}{\multirow{2}{*}{\textbf{Knowledge (French)}}}        & Le rapport donne également un aperçu de la discrimination à laquelle sont confrontées les personnes intersexes                               \\
\multicolumn{1}{l|}{}                                               & et   transgenres dans le domaine de l'emploi, ainsi que des niveaux de   harcèlement, de violence et de préjugés.     \\ \midrule
\multicolumn{2}{c}{\textbf{Source and Reference}} \\ \midrule
\multicolumn{1}{l|}{\multirow{2}{*}{\textbf{Source (German)}}}        & Der Bericht gibt auch einen Überblick über die Diskriminierung, der sich intersexuelle und Transgender-Personen                                     \\
\multicolumn{1}{l|}{}                                               &  im Berufsleben   ausgesetzt sehen sowie über das Ausmaß   von Belästigung, Gewalt und Vorurteilskriminalität. \\ \midrule
\multicolumn{1}{l|}{\multirow{2}{*}{\textbf{Reference}}}            & \textcolor{blue}{The} report also \textcolor{blue}{gives} an overview of the discrimination faced by intersex and \textcolor{blue}{transgender} people                                 \\
\multicolumn{1}{l|}{}                                               & in the realm of \textcolor{blue}{employment,} as well as levels of harassment, violence and bias crimes.                                         \\ \midrule
\multicolumn{2}{c}{\textbf{Translation}} \\ \midrule
\multicolumn{1}{l|}{\multirow{2}{*}{\textsc{Prompt-tuning}}}        & \textcolor{blue}{The} report also \textcolor{blue}{gives} an overview of the discrimination to which inter-sexual and \textcolor{blue}{transgender} people                             \\
\multicolumn{1}{l|}{}                                               & are subjected in   their \textcolor{red}{professional life} and the extent of harassment, violence and prejudice   \textcolor{red}{crime}.                         \\ \midrule
\multicolumn{1}{l|}{\textsc{Prefix-tuning}}                         & \textcolor{blue}{The} \textcolor{red}{subject of the report is the subject of the report}.                                                                        \\ \midrule
\multicolumn{1}{l|}{\multirow{2}{*}{\textsc{LoRA}}}                 & \textcolor{blue}{The} report also \textcolor{blue}{gives} an overview of the discrimination faced by inter-sexual and \textcolor{blue}{transgender} people                             \\
\multicolumn{1}{l|}{}                                               & in their \textcolor{red}{working   lives} and the extent of harassment, violence and prejudice.                                                   \\ \midrule
\multicolumn{1}{l|}{\multirow{2}{*}{\textsc{Parallel Adapter}}}     & \textcolor{blue}{The} report also \textcolor{blue}{gives} an overview of the \textcolor{red}{present} discrimination faced by inter-sexual and \textcolor{blue}{transgender} people                     \\
\multicolumn{1}{l|}{}                                               & in the \textcolor{blue}{workplace}, as well as the extent of harassment, violence and prejudice.                                                 \\ \midrule
\multicolumn{1}{l|}{\multirow{2}{*}{\textsc{Prompt Gating} (\textit{Ours})}} & \textcolor{blue}{The} report also \textcolor{blue}{gives} an overview of the discrimination suffered by inter-sexual and \textcolor{blue}{transgender} people                          \\
\multicolumn{1}{l|}{}                                               & in \textcolor{blue}{employment},   as well as the extent of harassment, violence and prejudice.                                                    \\ \bottomrule
\toprule
\multicolumn{2}{c}{\textit{Example 3}} \\ \midrule
\multicolumn{2}{c}{\textbf{Constraint}} \\ \midrule
\multicolumn{1}{l|}{\textbf{Keywords}}          & "\textcolor{blue}{attempt}"                                                       \\ \midrule
\multicolumn{1}{l|}{\textbf{Tense}}                & The tense of this sentence is the \textcolor{blue}{future}   tense.             \\ \midrule
\multicolumn{1}{l|}{\textbf{Knowledge (French)}}        & Demain, il tentera de s'entraîner avec   l'équipe.            \\ \midrule
\multicolumn{2}{c}{\textbf{Source and Reference}} \\ \midrule
\multicolumn{1}{l|}{\textbf{Source (German)}}        & Morgen wird er   versuchen, mit der Mannschaft zu trainieren. \\ \midrule
\multicolumn{1}{l|}{\textbf{Reference}}            & Tomorrow he \textcolor{blue}{will attempt} to train with   the team.            \\ \midrule
\multicolumn{2}{c}{\textbf{Translation}} \\ \midrule
\multicolumn{1}{l|}{\textsc{Prefix-tuning}}        & \textcolor{red}{This is the subject of this article.}                        \\ \midrule
\multicolumn{1}{l|}{\textsc{Parallel Adapter}}     & Tomorrow he \textcolor{blue}{will} \textcolor{red}{try} to train with the   team.                \\ \midrule
\multicolumn{1}{l|}{\textsc{LoRA}}                 & The team he \textcolor{blue}{will} \textcolor{red}{try} to train with \textcolor{red}{the   future}.              \\ \midrule
\multicolumn{1}{l|}{\textsc{Prompt-tuning}}        & Tomorrow he \textcolor{blue}{will}   \textcolor{red}{try} to train with the team.                \\ \midrule
\multicolumn{1}{l|}{\textsc{Prompt Gating} (\textit{Ours})} & Tomorrow he \textcolor{blue}{will attempt} to train with   the team.            \\ \bottomrule
\end{tabular}
}
        
        \caption{\label{tab:main2_example} Examples of multi-aspect controllable machine translation. ``\textbf{Keywords}'' denotes the given keywords that should be included in the translation. ``\textbf{Tense}'' denotes the input indicating the tense of the translation results. Similarly, ``\textbf{Knowledge (French)}'' denotes the external knowledge (i.e., French synonymous sentence). Some translations that {\color{blue}satisfy} the constraints are highlighted in blue, while some translations that {\color{red}fail to satisfy} the constraints are highlighted in red.}
    \end{center}
\end{table*}
%

\end{CJK*}

\end{document}